%% file: MAIN.tex
\documentclass[sigconf]{acmart}
\usepackage[super]{nth}
\usepackage{amsmath}
\usepackage{amsthm}
\usepackage{bm}
\usepackage{array}
\usepackage{makecell}
\usepackage{multirow}
\usepackage{tabularx}
\usepackage{paralist}
\usepackage{graphicx}
\usepackage{caption}
\usepackage{subcaption}
\usepackage{url}
\usepackage{xcolor}
\usepackage{enumitem}
\usepackage{float}
\usepackage[flushleft]{threeparttable}
\usepackage[ruled,vlined,linesnumbered]{algorithm2e}

\makeatletter
\newcommand{\mathleft}{\@fleqntrue\@mathmargin0pt}

\newtheorem{thm}{Theorem}

\theoremstyle{definition}

\newcounter{todocounter}
\DeclareMathOperator*{\argmin}{arg\,min}

\AtBeginDocument{%
  \providecommand\BibTeX{{%
    \normalfont B\kern-0.5em{\scshape i\kern-0.25em b}\kern-0.8em\TeX}}}

\copyrightyear{2022} 
\acmYear{2022} 
\setcopyright{acmcopyright}\acmConference[KDD '22]{Proceedings of the 28th ACM SIGKDD Conference on Knowledge Discovery and Data Mining}{August 14--18, 2022}{Washington, DC, USA}
\acmBooktitle{Proceedings of the 28th ACM SIGKDD Conference on Knowledge Discovery and Data Mining (KDD '22), August 14--18, 2022, Washington, DC, USA}
\acmPrice{15.00}
\acmDOI{10.1145/3534678.3539404}
\acmISBN{978-1-4503-9385-0/22/08}

\begin{document}
\setlength{\abovedisplayskip}{3.75pt plus 0pt}%
\setlength{\belowdisplayskip}{3.75pt plus 0pt}%
\title{Improving Fairness in Graph Neural Networks \\via Mitigating Sensitive Attribute Leakage}
\renewcommand{\shorttitle}{Improving Fairness in Graph Neural Networks via Mitigating Sensitive Attribute Leakage}

\author{Yu Wang}
\email{yu.wang.1@vanderbilt.edu}
\affiliation{%
  \institution{Vanderbilt University}
  \country{}
}

\author{Yuying Zhao}
\email{yuying.zhao@vanderbilt.edu}
\affiliation{%
  \institution{Vanderbilt University}
  \country{}
}

\author{Yushun Dong}
\email{yd6eb@virginia.edu}
\affiliation{%
  \institution{University of Virginia}
  \country{}
}

\author{Huiyuan Chen}
\email{hxc501@case.edu}
\affiliation{%
  \institution{Case Western Reserve University}
  \country{}
}

\author{Jundong Li}
\email{jundong@virginia.edu}
\affiliation{%
  \institution{University of Virginia}
  \country{}
}

\author{Tyler Derr}
\email{tyler.derr@vanderbilt.edu}
\affiliation{%
  \institution{Vanderbilt University}
  \country{}
}

\begin{abstract}
Graph Neural Networks (GNNs) have shown great power in learning node representations on graphs. However, they may inherit historical prejudices from training data, leading to discriminatory bias in predictions. Although some work has developed fair GNNs, most of them directly borrow fair representation learning techniques from non-graph domains without considering the potential problem of sensitive attribute leakage caused by feature propagation in GNNs. However, we empirically observe that feature propagation could vary the correlation of previously innocuous non-sensitive features to the sensitive ones. This can be viewed as a leakage of sensitive information which could further exacerbate discrimination in predictions. Thus, we design two feature masking strategies according to feature correlations to highlight the importance of considering feature propagation and correlation variation in alleviating discrimination. Motivated by our analysis, we propose Fair View Graph Neural Network (FairVGNN) to generate fair views of features by automatically identifying and masking sensitive-correlated features considering correlation variation after feature propagation. Given the learned fair views, we adaptively clamp weights of the encoder to avoid using sensitive-related features. Experiments on real-world datasets demonstrate that FairVGNN enjoys a better trade-off between model utility and fairness. Our code is publicly available at \href{https://github.com/YuWVandy/FairVGNN}{\textcolor{blue}{https://github.com/YuWVandy/FairVGNN}}.

\vspace{-2ex}
\end{abstract}

\keywords{\vspace{-3.5ex}\\
graph neural network, fairness, sensitive attribute leakage\vspace{-0.75ex}}

\ccsdesc[500]{Information systems~Data mining\vspace{-2.25ex}}

\maketitle

\input{introduction}
\input{preliminary}
\input{leakage}

\input{framework}
\input{experiments}
\input{relatedwork}

\input{conclusion}


\vspace{-2ex}
\bibliographystyle{ACM-Reference-Format}
\bibliography{references}









\appendix

\input{appendix}

\end{document}

%% file: introduction.tex
\vspace{-.9ex}
\section{Introduction}\label{sec-introduction}
As the world becomes more connected, graph mining is playing a crucial role in many domains such as drug discovery  and recommendation system~\cite{fan2019graph, chen2021structured, chemicalx}. As one of its major branches, learning 
informative node representation is a fundamental solution to many real-world problems 
such as node classification and link prediction~\cite{TDGNN, zhang2018link}. Numerous data-driven models have been developed for learning node representations, among which Graph Neural Networks (GNNs) have achieved unprecedented success owing to the combination of neural networks and feature propagation~\cite{GCN, APPNP, DAGNN}. 
Despite the significant progress of GNNs in 
capturing higher-order neighborhood information
~\cite{GCNII}, leveraging multi-hop dependencies~\cite{TDGNN}, 
and recognizing complex local topology contexts~\cite{wijesinghe2022a}, predictions of GNNs have been demonstrated to be unfair and perpetuate undesirable discrimination~\cite{dai2021say, shumovskaia2021linking, xu2021towards, nifty, bose2019compositional}.

Recent studies have revealed that historical data may include 
previous discriminatory decisions dominated by sensitive features~\cite{mehrabi2021survey,du2020fairness}. Thus, node representations learned from such data may explicitly inherit the existing societal biases and hence exhibit unfairness when applied in practice. Besides the sensitive features, network topology also serves as an implicit source of societal bias~\cite{EDITS, dai2021say}. By the principle of network homophily~\cite{mcpherson2001birds}, nodes with similar sensitive features tend to form closer connections than dissimilar ones. Since feature propagation smooths representations of neighboring nodes while separating distant ones, representations of nodes in different sensitive groups are further segregated and their corresponding predictions are unavoidably over-associated with sensitive features.

Besides above topology-induced bias, feature propagation could introduce another potential issue, termed as the sensitive information leakage. Since feature propagation naturally allows feature interactions among neighborhoods, the correlation between two feature channels is likely to vary after feature propagation, which is termed as correlation variation. As such, some original innocuous feature channels that have lower correlation to sensitive channels and encode less sensitive information may become highly correlated to sensitive ones after feature propagation and hence encode more sensitive information, which is termed as {\it sensitive attribute leakage}.
Some research efforts have been invested in alleviating discrimination  made by GNNs. However, they either borrow approaches from traditional fair representation learning such as adversarial debiasing~\cite{dai2021say, bose2019compositional} and contrastive learning~\cite{kose2021fairness} or directly debiasing node features and graph topology~\cite{EDITS, nifty} while overlooking the sensitive attribute leakage caused by correlation variation. 

\indent 
In this work, we study a novel and detrimental phenomenon where feature propagation can vary feature correlations and cause the leakage of sensitive information to innocuous features. To address this issue, we propose a principled framework Fair View Graph Neural Network (FairVGNN) to effectively learn fair node representations and avoid sensitive attribute leakage. Our major contributions are as follows:
\vspace{-1ex}
\begin{itemize}[leftmargin=*]
    \item \textbf{Problem}: We investigate the novel phenomenon that feature propagation could vary feature correlations and cause sensitive attribute leakage to innocuous feature channels, which could further exacerbate discrimination in predictions. 
    
    \item \textbf{Algorithm}: To prevent sensitive attribute leakage, we propose a novel framework FairVGNN to automatically learn fair views by identifying and masking sensitive-correlated channels and adaptively clamping weights to avoid leveraging sensitive-related features in learning fair node representations.
    
    \item \textbf{Evaluation}: We perform experiments on real-world datasets to corroborate that FairVGNN can approximate the model utility while reducing discrimination.
\end{itemize}

\indent Section~\ref{sec-preliminary} introduces preliminaries. In Section~\ref{sec-leakage}, we formally introduce the phenomenon of correlation variation and sensitive attribute leakage in GNNs and design two feature masking strategies to highlight the importance of circumventing sensitive attribute leakage for alleviating discrimination. To automatically identify/mask sensitive-relevant 
features, we propose FairVGNN in Section~\ref{sec-method}, which consists of a generative adversarial debiasing module to prevent sensitive attribute leakage from the input perspective by learning fair feature views and an adaptive weight clamping module to prevent sensitive attribute leakage from the model perspective by clamping weights of sensitive-correlated channels of the encoder. In Section~\ref{sec-experiment}, we evaluate FairVGNN by performing extensive experiments. Related work is presented in Section~\ref{sec-relatedwork}. Finally, we conclude and discuss future work in Section~\ref{sec-conclusion}. 


%% file: preliminary.tex
\vspace{-1.25ex}
\section{Preliminaries}\label{sec-preliminary}

\subsection{Notations}
We denote an attributed graph by $G = (\mathcal{V}, \mathcal{E}, \mathbf{X}, \mathbf{A})$ where $\mathcal{V} = \{v_1, ..., v_n\}$ is the set of $n$ nodes with $\mathbf{Y}\in\mathbb{R}^{n}$ specifying their labels, $\mathcal{E}$ is the set of $m$ edges with $e_{ij}$ being the edge between nodes $v_i$ and $v_j$, and $\mathbf{X} \in \mathbb{R}^{n\times d}$ is the node feature matrix with $\mathbf{X}_{i} = \mathbf{X}[i, :]^{\top} \in \mathbb{R}^d$ indicating the features of node $v_i$, $\mathbf{X}_{:j} = \mathbf{X}[:, j]\in\mathbb{R}^n$ indicating the $j^{\text{th}}$-channel feature. The network topology is described by its adjacency matrix $\mathbf{A}\in \{0, 1\}^{n\times n}$, where $\mathbf{A}_{ij} = 1$ 
when $e_{ij} \in \mathcal{E}$,
and $\mathbf{A}_{ij} = 0$ otherwise. Node sensitive features are specified by the $s^{\text{th}}$-channel of $\mathbf{X}$, i.e., $\mathbf{S} = \mathbf{X}_{:s} \in \mathbb{R}^{n}$. Details of all notations used in this work are summarized in Table~\ref{tb:symbols} in Appendix~\ref{sec-notation}.

\vspace{-1.5ex}
\subsection{Fairness in Machine Learning}
Group fairness and individual fairness are two commonly encountered fairness notions in real life~\cite{du2020fairness}. Group fairness emphasizes that algorithms should not yield discriminatory outcomes for any specific demographic group~\cite{EDITS} while individual fairness requires that similar individuals be treated similarly~\cite{dong2021individual}. Here we focus on group fairness with a binary sensitive feature, i.e., $\mathbf{S} \in \{0, 1\}^n$, but our framework could be generalized to multi-sensitive groups and we leave this as one future direction. 
Following~\cite{dai2021say, nifty, EDITS}, we employ the difference of statistical parity and equal opportunity between two different sensitive groups, to evaluate the model fairness:
\begin{equation}\label{eq-spdelta}
    \Delta_{\text{sp}} = |P(\hat{y} = 1|s = 0) - P(\hat{y} = 1|s = 1)|,
\end{equation}
\vskip -3ex
\begin{equation}
\vspace{-.5ex}
    \Delta_{\text{eo}} = |P(\hat{y} = 1|y = 1, s = 0) - P(\hat{y} = 1 |y = 1, s = 1)|,
\end{equation}
where $\Delta_{\text{sp}} (\Delta_{\text{eo}})$ measures the difference of the independence level of the prediction $\hat{y}$ (true positive rate) on the sensitive feature $s$ between two groups. Since group fairness expects algorithms to yield similar outcomes for different demographic groups, fairer machine learning models seek lower $\Delta_{\text{sp}}$ and $\Delta_{\text{eo}}$.

\begin{figure*}[t]
     \centering
     \includegraphics[width=1.0\textwidth]{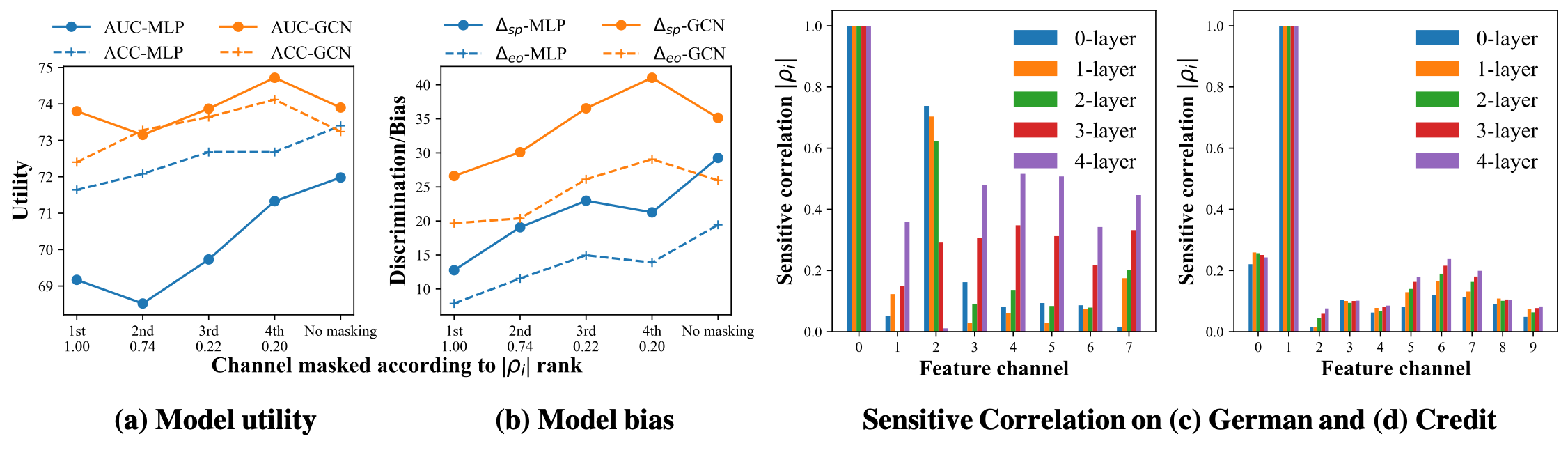}
     \vskip -3.75ex
     \caption{Initial empirical investigation on sensitive leakage and correlation variation on German dataset. (a)-(b) visualize the relationships between model utility/fairness and the sensitive correlation $\boldsymbol{\rho}_i$ of each masked feature channel\protect\footnotemark. Masking channel with less sensitive correlation leads to more biased predictions and sometimes higher model utility. (c)-(d) shows the correlation variation caused by feature propagation on German and Credit datasets. In (c), we can see sensitive correlations of the 2$^{\text{nd}}$ and 7$^{\text{th}}$ feature channel significantly change after propagation while in (d), the correlations do not change so much.}
     \label{fig-prelim}
     \vskip -1.75ex
\end{figure*}

%% file: leakage.tex
\vspace{-1.7ex}
\section{Sensitive attribute leakage and correlation variation}\label{sec-leakage}
In this section, we 
study the phenomenon where sensitive information leaks to innocuous feature channels after their correlations to the sensitive feature increase during feature propagation in GNNs, which we define as {\it sensitive attribute leakage}
. We first empirically verify 
feature channels with higher correlation to the sensitive channel would cause more discrimination in predictions~\cite{zhao2021you}. We denote the Pearson correlation coefficient of the $i^{\text{th}}$-feature channel to the sensitive channel as sensitive correlation and compute it as:
\begin{equation}\label{eq-lc}
    \boldsymbol{\rho}_{i} = \frac{\mathbb{E}_{v_j\sim\mathcal{V}}\big((\mathbf{X}_{ji} - \mu_{i})(\mathbf{S}_j - \mu_{s})\big)}{\sigma_{i}\sigma_{s}}, \forall i\in \{1, 2, ..., d\},
\end{equation}
where $\mu_i, \sigma_i$ denote the mean and standard deviation of the channel $\mathbf{X}_{:i}$. Intuitively, higher $\boldsymbol{\rho}_i$ indicates that the $i^{\text{th}}$-feature channel encodes more sensitive-related information, which would impose more discrimination in the prediction. To further verify this assumption, we mask each channel and train a 1-layer MLP/GCN followed by a linear layer to make predictions. As suggested by~\cite{nifty}, we do not add any activation function in the MLP/GCN to avoid capturing any nonlinearity.


Figure~\ref{fig-prelim}(a)-(b) visualize the relationships between the model utility/bias and the sensitive correlation of each masked feature channel. Clearly, we see that the discrimination does still exist even though we mask the sensitive channel (1$^\text{st}$). 
Compared with no masking situation, $\Delta_{\text{sp}}$ and $\Delta_{\text{eo}}$ almost always become lower when we mask 
other non-sensitive feature channels (2$^\text{nd}$-4$^\text{th}$), which indicates the leakage of sensitive information to other non-sensitive feature channels. Moreover, we observe the decreasing trend of $\Delta_{\text{sp}}$ and $\Delta_{\text{eo}}$ when masking channels with higher sensitive correlation since these channels encode more sensitive information and masking them would alleviate more discrimination. 

Following the above observation, one natural way to prevent sensitive attribute leakage and alleviate discrimination is to mask the sensitive features as well as its highly-correlated non-sensitive features. However, feature propagation in GNNs could change feature distributions of different channels and consequentially vary feature correlations as shown by Figure~\ref{fig-prelim}(c) where we visualize the sensitive correlations of the first 8 feature channels on German after a certain number of propagations. We see that correlations between the sensitive features 
and other channels change 
after propagation. For example, some feature channels that are originally irrelevant to the sensitive one, such as the $7^{\text{th}}$ feature channel, 
become highly-correlated and hence encode more sensitive information.

\footnotetext{We respectively mask each feature channel and train a 1-layer MLP/GCN followed by a linear prediction layer. Dataset and experimental details are given in Section~\ref{sec-experimentsetting}.}

\begin{table}[t]
\footnotesize
\setlength{\extrarowheight}{.12pt}
\setlength\tabcolsep{3.2pt}
\centering
\caption{Evaluating model utility and fairness when using various strategies of feature masking (or no masking).}
\vskip -2ex
\begin{tabular}{ll|llll|llll}
\hline
\multirow{2}{*}{\textbf{Encoder}} & \multirow{2}{*}{\textbf{Strategy}} & \multicolumn{4}{c|}{\textbf{German}} & \multicolumn{4}{c}{\textbf{Credit}} \\ 
 &  & AUC & F1 & $\Delta_{\text{sp}}$ & $\Delta_{\text{eo}}$ & AUC & F1 & $\Delta_{\text{sp}}$ & $\Delta_{\text{eo}}$ \\
 \hline
\multirow{3}{*}{\textbf{MLP}} & S$_0$ & 71.98 & 82.32 & 29.26 & 19.43 & 74.46 & 81.64 & 11.85 & 9.61 \\
 & S$_1$ & 69.89 & 81.37 & 8.25 & 4.75 & 73.49 & 81.50 & 11.50 & 9.20 \\
 & S$_2$ & 70.54 & 81.44 & 6.58 & 3.24 & 73.49 & 81.50 & 11.50 & 9.20 \\
 \hline
\multirow{3}{*}{\textbf{GCN}} & S$_0$ & 74.11 & 82.46 & 35.17 & 25.17 & 73.86 & 81.92 & 12.86 & 10.63 \\
 & S$_1$ & 73.78 & 81.65 & 11.39 & 9.60 & 72.92 & 81.84 & 12.00 & 9.70 \\
 & S$_2$ & 72.75 & 81.70 & 8.29 & 6.91 & 72.92 & 81.84 & 12.00 & 9.70\\
 \hline
 \multirow{3}{*}{\textbf{GIN}} & S$_0$ & 72.71 & 82.78 & 13.56 & 9.47 & 74.36 & 82.28 & 14.48 & 12.35 \\
 & S$_1$ & 71.66 & 82.50 & 3.01 & 1.72 & 73.44 & 83.23 & 14.29 & 11.79\\
 & S$_2$ & 70.77 & 83.53 & 1.46 & 2.67 & 73.28 & 83.27 & 13.96 & 11.34\\
 \hline
 
\end{tabular}
\begin{tablenotes}
  \footnotesize
  \item \textbf{*} S$_0$: training using the original feature matrix $\mathbf{X}$ without any masking.
  
  \item \textbf{*} S$_1$/S$_2$: training with masking the top-$4$ channels based on the rank of $\boldsymbol{\rho}^{\text{origin}}$/$\boldsymbol{\rho}^{\text{prop}}$.
\end{tablenotes}
\vskip -2ex
\vspace{-4ex}
\label{tab-prelim}
\end{table}

After observing that feature propagation could vary feature correlation and cause sensitive attribute leakage,
we devise two simple but effective masking strategies to highlight the importance of considering correlation variation and sensitive attribute leakage in alleviating discrimination.
Specifically, we first compute sensitive correlations of each feature channel according to 1) the original features $\boldsymbol{\rho}^{\text{origin}}$ and 2) the propagated features $\boldsymbol{\rho}^{\text{prop}}$. Then, we manually mask top-$k$ feature channels according to the absolute values of correlation given by $\boldsymbol{\rho}^{\text{origin}}$ and $\boldsymbol{\rho}^{\text{prop}}$, respectively, and train MLP/GCN/GIN on German/Credit dataset shown in Table~\ref{tab-prelim}. 
Detailed experimental settings are presented in Section~\ref{sec-experiment}. From Table~\ref{tab-prelim},
we have following insightful observations:
\begin{inparaenum}
    \item Within the same encoder, masking sensitive and its related feature channels (S$_1$, S$_2$) would alleviate the discrimination while downgrading the model utility compared with no-masking (S$_0$).
    \item GCN achieves better model utility but causes more bias compared with MLP on German and Credit. This implies graph structures also encode bias and leveraging them could aggravate discrimination in predictions, which is consistent with recent work~\cite{dai2021say, EDITS}.
    \item Most importantly, S$_2$ achieves lower $\Delta_{\text{sp}}, \Delta_{\text{eo}}$ than S$_1$ for both MLP and GCN on German because the rank of sensitive correlation changes after feature propagation and masking according to S$_2$ leads to better fairness, which highlights the importance of considering feature propagation in determining which feature channels are more sensitive-correlated and required to be masked. Applying S$_1$ achieves the same utility/bias as S$_2$ on Credit due to less 
    correlation variations shown in Figure~\ref{fig-prelim}(d).
\end{inparaenum} 

To this end, we argue that it is necessary to consider feature propagation in masking feature channels in order to alleviate discrimination.  However, the correlation variation heavily depends on the propagation mechanism of GNNs.
To tackle this challenge, we formulate our problem as:

\textit{Given an attributed network $\mathcal{G} = (\mathcal{V}, \mathcal{E}, \mathbf{X}, \mathbf{A})$ with labels $\mathbf{Y}$ for a subset of nodes $\mathcal{V}_l \subset \mathcal{V}$, we aim to learn a fair view generator $g_{\bm{\Theta}_g}: g_{\bm{\Theta}_g}(\mathbf{X}) \rightarrow \widetilde{\mathbf{X}}$ with the expectation of simultaneously preserving task-related information and discarding sensitive information such that the downstream node classifier $f_{\bm{\Theta}_f}: f_{\bm{\Theta}_f}(\mathbf{A}, \widetilde{\mathbf{X}}) \rightarrow \mathbf{Y}$ trained on $\widetilde{\mathbf{X}}$ could achieve better trade-off between model utility and fairness.}


%% file: framework.tex
\vspace{-1ex}
\section{Framework}\label{sec-method}
\begin{figure}[t!]
     \centering
     \includegraphics[width=.5\textwidth]{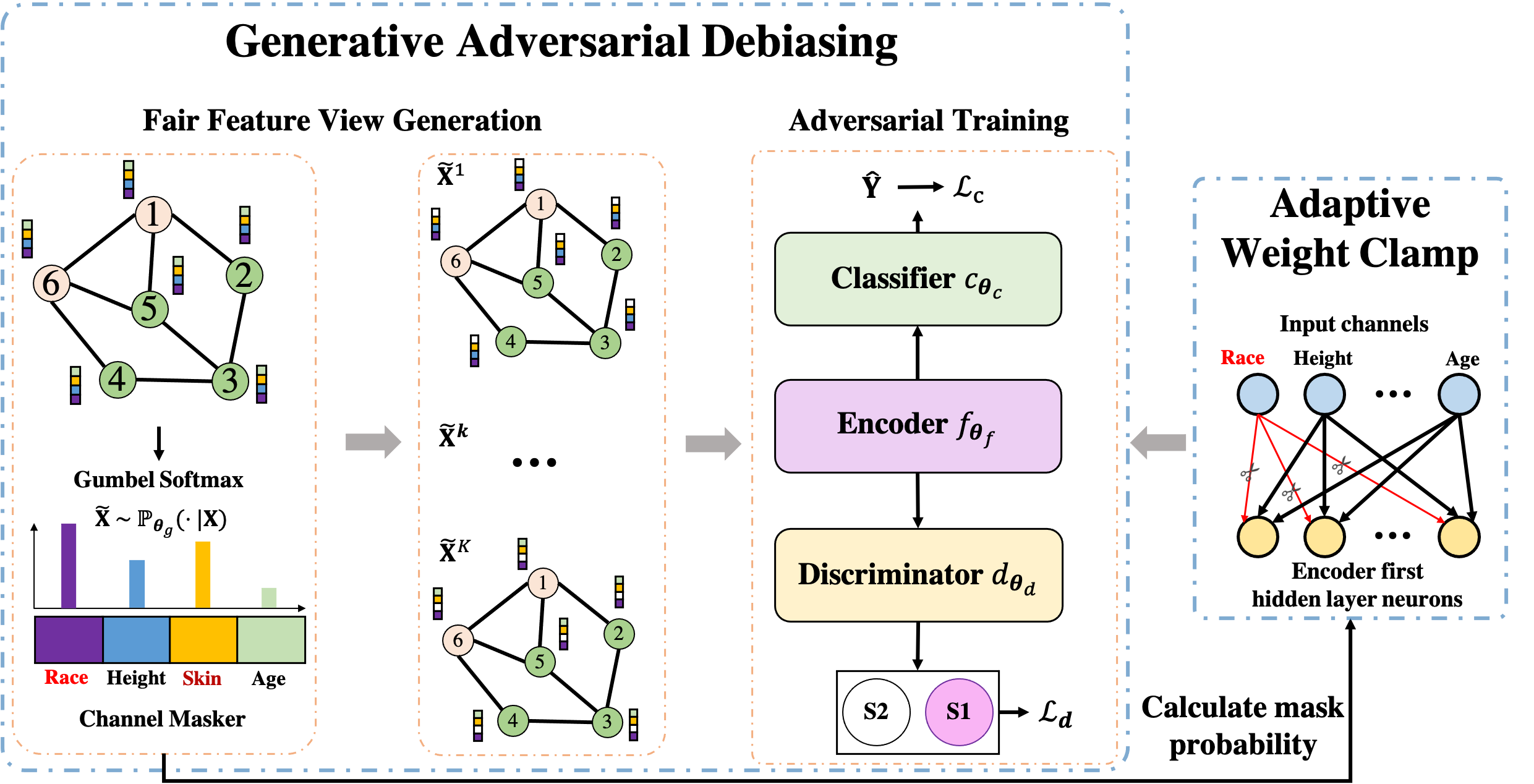}
     \caption{An overview of the Fair View Graph Neural Network (FairVGNN), with two main modules: (a) generative adversarial debiasing to learn fair view of features and (b) adaptive weight clamping to clamp weights of sensitive-related channels of the encoder.}
     \label{fig-fairvgnn}
\end{figure}

In this section, we give a detailed description of FairVGNN (shown in Figure~\ref{fig-fairvgnn}), which includes the generative adversarial debiasing module and the adaptive weight clamping module. In the first module, we learn a generator that  generates different fair views of features to obfuscate the sensitive discriminator such that the encoder could obtain fair node representations for downstream tasks. In the second module, we propose to clamp weights of the encoder based on learned fair feature views, and provide a theoretical justification on its equivalence to minimizing the upper bound of the difference of representations between two different sensitive groups. Next, we introduce the details of each component.

\vspace{-1.5ex}
\subsection{Generative Adversarial Debiasing}
This module includes a fair view generator $g_{\bm{\Theta}_g}$, a GNN-based encoder $f_{\bm{\Theta}_f}$, a sensitive discriminator $d_{\bm{\Theta}_d}$, and a classifier $c_{\bm{\Theta}_c}$ parametrized by $\bm{\Theta}_g, \bm{\Theta}_f, \bm{\Theta}_d, \bm{\Theta}_c$, respectively.
We assume the view generator $g_{\bm{\Theta}_g}$ to be a learnable latent distribution from which we sample $K$-different masks and generate $K$-corresponding views $\widetilde{\mathbf{X}}^k, k\in\{1,2,..., K\}$. The latent distribution would be updated towards generating less-biased views $\widetilde{\mathbf{X}}$ and the stochasticity of each view would enhance the model generalizability. Then each of these $K$-different views $\widetilde{\mathbf{X}}^k$ are fed to the encoder $f_{\bm{\Theta}_f}$ together with the network topology $\mathbf{A}$ to learn node representations $\widetilde{\mathbf{H}}^k$ for downstream classifier $c_{\bm{\Theta}_c}$. Meanwhile, the learned node representations $\widetilde{\mathbf{H}}^k$ are used by the sensitive discriminator $d_{\bm{\Theta}_d}$ to predict nodes' sensitive features. This paves us a way to adopt adversarial learning to obtain the optimal fair view generator $g_{\bm{\Theta}_g^{*}}$ 
where the generated views encode as much task-relevant information while discarding as much bias-relevant information as possible.
We begin with introducing the fairness-aware view generator $g_{\bm{\Theta}_g}$.

\vspace{-1ex}
\subsubsection{Fairness-aware View Generator}\label{sec-g}
As observed in Table~\ref{tab-prelim}, discrimination could be traced back to the sensitive features as well as their 
highly-correlated non-sensitive features.
Therefore, we propose to learn a view generator that automatically identifies and masks these features.
More specifically, assuming the view generator as a conditional distribution $\mathbb{P}_{\widetilde{G}}$ parametrized by $\bm{\Theta}_g$, since bias originates from the node features $\mathbf{X}$ and is further varied by the graph topology $\mathbf{A}$, the conditional distribution of the view generator can be further expressed as a joint distribution of the attribute generator and the topological generator as $\mathbb{P}_{\widetilde{G}} = \mathbb{P}_{\widetilde{\mathbf{X}}, \widetilde{\mathbf{A}}}$. Since  our sensitive discriminator $d_{\bm{\Theta}_d}$ is directly trained on the learned node representations from GNN-based encoder $f_{\bm{\Theta}_f}$ as described in Section~\ref{sec-cd}, we already consider the proximity-induced bias in alleviating discrimination and hence the network topology  is assumed to be fixed here, i.e., $\mathbb{P}^{\bm{\Theta}_g}_{\widetilde{\mathbf{X}}, \widetilde{\mathbf{A}}} =\mathbb{P}^{\bm{\Theta}_g}_{\widetilde{\mathbf{X}}}$. We will leave the joint generation of fair feature and topological views as one future work.

Instead of generating $\widetilde{\mathbf{X}}$ from scratch that completely loses critical information for GNN predictions, we generate $\widetilde{\mathbf{X}}$ conditioned on the original node features $\mathbf{X}$, i.e., $\mathbb{P}^{\bm{\Theta}_g}_{\widetilde{\mathbf{X}}} = \mathbb{P}^{\bm{\Theta}_g}_{\widetilde{\mathbf{X}}}(\cdot|\mathbf{X})$. Following the preliminary experiments, we model the generation process of $\widetilde{\mathbf{X}}$ as identifying and masking sensitive features and their highly-correlated features in $\mathbf{X}$. One natural way is to select features according to their correlations $\boldsymbol{\rho}_{i}$ to the sensitive features $\mathbf{S}$ as defined in Eq.~\eqref{eq-lc}. However, as shown by Figure~\ref{fig-prelim}(c), feature propagation in GNNs triggers the correlation variation. 
Thus, instead of masking according to initial correlations that might change after feature propagation, 
we train a learnable mask for feature selections in a data-driven fashion. Denote our mask as $\mathbf{m} = [m_1, m_2, ..., m_d]\in\{0, 1\}^d$ 
so that:
\begin{equation}
    \widetilde{\mathbf{X}} = \mathbf{X}\odot\mathbf{m} = [\mathbf{X}_{1}^{\top}\odot \mathbf{m}, \mathbf{X}_{2}^{\top}\odot \mathbf{m}, ..., \mathbf{X}_{n}^{\top}\odot \mathbf{m}],
\end{equation}
then learning the conditional distribution of the feature generator $\mathbb{P}^{\bm{\Theta}_g}_{\widetilde{\mathbf{X}}}(\cdot|\mathbf{X})$ is transformed to learning a sampling distribution of the masker $\mathbb{P}^{\bm{\Theta}_g}_{\mathbf{m}}$. We assume the probability of masking each feature channel independently follows a Bernoulli distribution, i.e., $m_i \sim \text{Bernoulli}(1 - p_i), \forall i\in \{1, 2, ..., d\}$ with each feature channel $i$ being masked with the learnable probability $p_i\in\mathbb{R}$. In this way, we can learn which feature channels should be masked to achieve less discrimination through gradient-based techniques.
Since the generator $g_{\bm{\Theta}_g}$ aims to obfuscate the discriminator $d_{\bm{\Theta}_d}$ that predicts the sensitive features based on the already-propagated node representations $\widetilde{\mathbf{H}}$ from the encoder $f_{\bm{\Theta}_f}$, the generated fair feature view $\widetilde{\mathbf{X}}$ would consider the effect of correlation variation by feature propagation rather than blindly follow the order of the sensitive correlations computed by the original features $\mathbf{X}$. Generating fair feature view $\widetilde{\mathbf{X}}$ and forwarding it through the encoder $f_{\bm{\Theta}_f}$ and the classifier $c_{\bm{\Theta}_c}$ to make predictions involve sampling masks $\mathbf{m}$ from the categorical Bernoulli distribution, the whole process of which is non-differentiable
due to the discreteness of masks. Therefore, we apply Gumbel-Softmax trick~\cite{jang2016categorical} to approximate the categorical Bernoulli distribution. Assuming for each channel $i$, we have a learnable sampling score $\boldsymbol{\pi}_i = [\pi_{i1}, \pi_{i2}]$ with $\pi_{i1}$ score keeping while $\pi_{i2}$ score masking the channel $i$. Then the categorical distribution $\text{Bernoulli}(1 - p_i)$ is softened by\footnote{We use $p_{i1}$ instead of $p_i$ thereafter to represent the probability of keeping channel $i$.}:
\begin{align}
    p_{ij} = \frac{\exp(\frac{\log(\pi_{ij}) + g_{ij}}{\tau})}{\sum_{k= 1}^2\exp(\frac{\log(\pi_{ik}) + g_{ik}}{\tau})}, \forall j = 1, 2, i\in\{1, 2, ..., d\},
\end{align}
\noindent where $g_{ij}\sim \text{Gumbel}(0, 1)$ and $\tau$ is the temperature factor controlling the sharpness of the Gumbel-Softmax distribution. Then, to generate $\widetilde{\mathbf{X}}$ after we sample masks $\mathbf{m}$ based on probability $p_{i1}$, we could either directly multiply feature channel $\mathbf{X}_{:i}$ by the probability $p_{i1}$ or solely append the gradient of $p_{i1}$ to the sampled hard mask\footnote{$\mathbf{m} =\mathbf{m} - p_{i1}.\text{detach()} + p_{i1}$}, both of which are differentiable and can be trained end to end. After we approximate the generator $g_{\bm{\Theta}_g}$ via Gumbel-Softmax, we next model the GNN-based encoder $f_{\bm{\Theta}_f}$ to capture the information of both node features $\mathbf{X}$ and network topology $\mathbf{A}$. 
\subsubsection{GNN-based Encoder}\label{sec-e}
In order to learn from both the graph topology and node features, we employ $L-$layer GNNs as our encoder-backbone to obtain node representations $\mathbf{H}^{L}$. Different graph convolutions adopt different propagation mechanisms,
resulting in different variations on feature correlations.
Here we select GCN~\cite{GCN}, GraphSAGE~\cite{Graphsage}, and GIN~\cite{GIN} as our encoder-backbones. In order to consider the variation induced by the propagation of GNN-based encoders, we apply the discriminator $d_{\bm{\Theta}_d}$ and classifier $c_{\bm{\Theta}_c}$ on top of the obtained node representations $\mathbf{H}^L$ from the GNN-based encoders. Since both of the classifier and the discriminator are to make predictions, one towards sensitive groups and the other towards class labels, their model architectures are similar and therefore we introduce them together next. 

\subsubsection{Classifier and Discriminator}\label{sec-cd}
Given node representations $\mathbf{H}^{L}$ obtained from any $L-$layer GNN-based encoder $f_{\bm{\Theta}_f}$, the classifier $c_{\bm{\Theta}_c}$ and the discriminator $d_{\bm{\Theta}_d}$ predict node labels $\hat{\mathbf{Y}}$ and sensitive attributes $\hat{\mathbf{S}}$ as:
\begin{equation}
\small
    \hat{\mathbf{Y}} = c_{\bm{\Theta}_c}(\mathbf{H}^L) = \sigma\big(\text{MLP}_c(\mathbf{H}^L)\big),~~~\hat{\mathbf{S}} = d_{\bm{\Theta}_d}(\mathbf{H}^L) = \sigma\big(\text{MLP}_d(\mathbf{H}^L)\big),
\end{equation}
where we use two different multilayer perceptrons (MLPs): $\mathbb{R}^{d^L}\rightarrow \mathbb{R}$ for the classifier and the discriminator, and $\sigma$ is the sigmoid operation.
After introducing the fairness-aware view generator, the GNN-based encoder, the MLP-based classifier and discriminator, we collect them together and adversarially train them with the following objective function. 
\vspace{-1ex}
\subsubsection{Adversarial Training}\label{sec-advtraining}
Our goal is to learn fair views from the original graph that encode as much task-relevant information while discarding as much sensitive-relevant information as possible. Therefore, we aim to optimize the whole framework from both the fairness and model utility perspectives. According to statistical parity, to optimize the fairness metric, a fair feature view should guarantee equivalent predictions between sensitive groups:
\begin{equation}\label{eq-minsp}
    \bm{\Theta}_g^{*} =\argmin_{\bm{\Theta}_g}\Delta_{\text{sp}} = \argmin_{\bm{\Theta}_g}|P(\hat{y} = 1| s = 0) - P(\hat{y} = 1 | s = 1)|,
\end{equation}
where $P(\hat{y}|s)$ is the predicted distribution given the sensitive feature. Assuming $\hat{y}$ and $s$ are conditionally independent given $\widetilde{\mathbf{H}}$~\cite{kamishima2011fairness}, to solve the global minimum of Eq.~\eqref{eq-minsp}, we leverage adversarial training
and compute the loss of the discriminator and generator $\mathcal{L}_d, \mathcal{L}_g$ as:

\vspace{-3ex}
\begin{equation}\label{eq-advd}
\small
\max_{\bm{\Theta}_d}\mathcal{L}_{\text{d}} = \mathbb{E}_{\widetilde{\mathbf{X}}\sim\mathbb{P}^{\bm{\Theta}_g}_{(\widetilde{\mathbf{X}}|\mathbf{X})}}
\mathbb{E}_{v_i\sim\mathcal{V}}
\bigg(\mathbf{S}_i\log\big(d_{\bm{\Theta}_d}(\widetilde{\mathbf{H}}^L_i)) + (1 - \mathbf{S}_i)\log(1 - d_{\bm{\Theta}_d}(\widetilde{\mathbf{H}}^L_i)\big)\bigg),
\end{equation}

\vspace{-3ex}
\begin{equation}\label{eq-advg}
\small
\min_{\bm{\Theta}_g}\mathcal{L}_{\text{g}} =\mathbb{E}_{\widetilde{\mathbf{X}}\sim\mathbb{P}^{\bm{\Theta}_g}_{(\widetilde{\mathbf{X}}|\mathbf{X})}}
\mathbb{E}_{v_i\sim\mathcal{V}}
\big(d_{\bm{\Theta}_d}(\widetilde{\mathbf{H}}^L_i) - 0.5\big)^2 + \alpha||\mathbf{m} - \mathbf{1}_d||_2^2,
\end{equation}
where $\widetilde{\mathbf{H}}^L_i = f_{\bm{\Theta}_f}(\widetilde{\mathbf{X}}_i, \mathbf{A})$ and $||\mathbf{m} - \mathbf{1}_d||_2^2$ regularizes the mask to be dense, which avoids masking out sensitive-uncorrelated but task-critical information. $\alpha$ is the hyperparamter. Intuitively, Eq.~\eqref{eq-advd} encourages our discriminator to correctly predict the sensitive features of each node under each generated view and Eq.~\eqref{eq-advg} requires our generator to generate fair feature views that enforce the well-trained discriminator to randomly guess the sensitive features.
In Theorem~\ref{thm-same}, we show that the  global minimum of Eq.~\eqref{eq-advd}-\eqref{eq-advg} is equivalent to the global minimum of Eq.~\eqref{eq-minsp}:
\vspace{-1ex}
\begin{thm}\label{thm-same}

Given $\widetilde{\mathbf{h}}^L$ as the representation of a specific node learned by L layer GNN-based encoder $f_{\bm{\Theta}_g}$ and $\alpha = 0$ in Eq.~\eqref{eq-advg}, the global optimum of Eq.~\eqref{eq-advd}-\eqref{eq-advg} is equivalent to the one of Eq.~\eqref{eq-minsp}.
\end{thm}
\vspace{-2ex}

\begin{proof}

Based on Proposition 1. in \cite{goodfellow2014generative} and Proposition 4.1. in \cite{dai2021say}, the optimal discriminator is $d_{\boldsymbol{\theta}_d^*}(\widetilde{\mathbf{h}}^L) = \frac{P(\widetilde{\mathbf{h}}^L|s = 1)}{P(\widetilde{\mathbf{h}}^L|s = 1) + P(\widetilde{\mathbf{h}}^L|s = 0)}$, which is exactly the probability when discriminator randomly guesses the sensitive features. Then we further substituted it into Eq.~\eqref{eq-advg} and the optimal generator is achieved when $d_{\boldsymbol{\theta}_d^*}(\widetilde{\mathbf{h}}^L) = 0.5, \text{i.e.}, P(\widetilde{\mathbf{h}}^L|s = 1) = P(\widetilde{\mathbf{h}}^L|s = 0)$. Then we have:
\begin{align}
    P(\hat{y} = 1|s &= 1) = \int_{\widetilde{\mathbf{h}}^L}P(\hat{y} = 1|\widetilde{\mathbf{h}}^L)P(\widetilde{\mathbf{h}}^L|s = 1)d\widetilde{\mathbf{h}}^L
    \nonumber\\& = \int_{\widetilde{\mathbf{h}}^L}P(\hat{y} = 1|\widetilde{\mathbf{h}}^L)P(\widetilde{\mathbf{h}}^L|s = 0)d\widetilde{\mathbf{h}}^L = P(\hat{y} = 1|s = 0), \nonumber
\end{align}
which is obviously the global minimum of Eq.~\eqref{eq-minsp}.
\end{proof}
\vspace{-2ex}
Note that node representations $\widetilde{\mathbf{H}}^L$ have already been propagated in GNN-based encoder $f_{\boldsymbol{\theta}_f}$ and therefore, the optimal discriminator $d_{\boldsymbol{\theta}_d^*}$  could identify sensitive-related features after correlation variation. Besides the adversarial training loss to ensure the fairness of the generated view, the classification loss for training the classifier $c_{\boldsymbol{\theta}_c}$ is used to guarantee the model utility:
\vspace{-2ex}
\begin{equation}
\small
\min_{\boldsymbol{\theta}_c}\mathcal{L}_{\text{c}} = -\mathbb{E}_{\widetilde{\mathbf{X}}\sim\mathbb{P}^{\boldsymbol{\theta}_g}_{(\widetilde{\mathbf{X}}|\mathbf{X})}}
\mathbb{E}_{v_i\sim\mathcal{V}}
\bigg(\mathbf{Y}_i\log\big(c_{\boldsymbol{\theta}_c}(\widetilde{\mathbf{H}}^L_i)\big) + (1 - \mathbf{Y}_i)\log\big(1 - c_{\boldsymbol{\theta}_c}(\widetilde{\mathbf{H}}^L_i)\big)\bigg)
\end{equation}


\vspace{-2ex}
\subsection{Adaptive Weight Clamping}\label{sec-wc}
Although the generator is theoretically guaranteed to achieve its global minimum by applying adversarial training,
in practice the generated views may still encode sensitive information and the corresponding classifier may still make discriminatory decisions. This is because of the unstability of the training process of adversarial learning~\cite{goodfellow2014generative} and the entanglement with training classifier.

To alleviate the above issue, we propose to adaptively clamp weights of the encoder $f_{\bm{\Theta}_f}$ based on the learned masking probability distribution from the generator $g_{\bm{\Theta}_g}$. After adversarial training, only the sensitive and its highly-correlated features would have higher probability to be masked and therefore, declining their contributions in $\widetilde{\mathbf{H}}^L$ by clamping their corresponding weights in the encoder would discourage the encoder from capturing these features and hence alleviate the discrimination. Concretely, within each training epoch after the adversarial training, we compute the probability of keeping features $\mathbf{p}\in\mathbb{R}^{d}$ by sampling $K$ masks and calculate their mean $\mathbf{p} = \sum_{k = 1}^K\mathbf{m}^k$. Then assuming the weights of the first layer in the encoder $f_{\bm{\Theta}_f}$ is $\mathbf{W}^{f, 1}\in\mathbb{R}^{d_1\times d}$, we clamp it by:
\begin{equation}\label{eq-wc}
    \mathbf{W}^{f, 1}_{ij} = \begin{cases}
        \mathbf{W}_{ij}^{f, 1}, & |\mathbf{W}_{ij}^{f, 1}| \le \epsilon*\mathbf{p}_j\\
        \text{sign}(\mathbf{W}_{ij}^{f, 1}) * \epsilon*\mathbf{p}_j, & |\mathbf{W}_{ij}^{f, 1}| > \epsilon*\mathbf{p}_j
    \end{cases},
\end{equation}
where $\epsilon\in\mathbb{R}$ is a prefix cutting threshold selected by hyperparameter tuning and $\text{sign}:\mathbb {R} \to \{-1,0,1\}$ takes the sign of $\mathbf{W}_{ij}^{f, 1}$. Intuitively, feature channels masked with higher probability (remained with lower probability $\mathbf{p}_j$) would have lower threshold in weight clamping and hence their contributions to the representations $\widetilde{\mathbf{H}}^{L}$ are weakened.
Next, we theoretically rationalize this adaptive weight clamping by demonstrating its equivalence to minimizing the upper bound of the difference of representations between two sensitive groups:
\vspace{-2ex}
\begin{thm}\label{thm-cffairness}
Given a 1-layer GNN encoder $f_{\boldsymbol{\theta}_f}$ with row-normalized adjacency matrix $\mathbf{D}^{-1}\mathbf{A}$ as the PROP and weight matrix $\mathbf{W}^{f, 1}$ as TRAN and further assume that features of nodes from two sensitive groups in the network independently and identically follow two different Gaussian distributions, i.e., $\mathbf{X}^{s_1}\sim\mathcal{N}(\boldsymbol{\mu}^{s_1}, \boldsymbol{\Sigma}^{s_1}), \mathbf{X}^{s_2}\sim\mathcal{N}(\boldsymbol{\mu}^{s_2}, \boldsymbol{\Sigma}^{s_2})$, then the difference of representations $\mathbf{H}^{s_1} - \mathbf{H}^{s_2}$ also follows a Gaussian with the 2-norm of its mean $\boldsymbol\mu$ as:
\vspace{-1.5ex}
\begin{equation}\label{eq-bound}
\footnotesize
\begin{split}
    &||\boldsymbol{\mu}||_2 = ||(2\chi - 1)\mathbf{W}^{f, 1}\Delta\boldsymbol{\mu}||_2
    \le (2\chi - 1)\big(\sum_{i = 1}^{d_1}(\sum_{r\in\mathcal{S}}\epsilon\mathbf{p}_r\Delta\boldsymbol{\mu}_r + \sum_{k\in\mathcal{NS}}\epsilon\mathbf{p}_k\Delta\boldsymbol{\mu}_k)^2\big)^{0.5}
\end{split}
\end{equation}
\end{thm}
\vspace{-2ex}
\noindent where $\Delta\boldsymbol{\mu} = \boldsymbol{\mu}^{s_1} - \boldsymbol{\mu}^{s_2}\in\mathbb{R}^{d}$ and $\mathcal{S}, \mathcal{NS}$ denote the sensitive and non-sensitive features, and $\chi$ is the network homophily.

\begin{proof}
Substituting the row-normalized adjacency matrix $\mathbf{D}^{-1}(\mathbf{A}+\mathbf{I})$, we have $f_{\boldsymbol{\theta}_f}(\mathbf{X}) = \mathbf{W}^{f, 1}\mathbf{D}^{-1}(\mathbf{A} + \mathbf{I})\mathbf{X}$, for any pair of nodes coming from two different sensitive groups $v_i\in\mathcal{V}_0, v_j\in\mathcal{V}_1$, we have:
\begin{equation}\label{eq-dist0}
\begin{split}
    &f_{\boldsymbol{\theta}_f}(\mathbf{X}_i) - f_{\boldsymbol{\theta}_f}(\mathbf{X}_j)
    = \mathbf{W}^{f, 1}\big(\mathbf{D}^{-1}(\mathbf{A} + \mathbf{I})\mathbf{X}\big)_i - \mathbf{W}^{f, 1}\big(\mathbf{D}^{-1}(\mathbf{A} + \mathbf{I})\mathbf{X}\big)_j
    \\=& \mathbf{W}^{f, 1}(\frac{1}{d_i + 1}\sum_{v_p\in\mathcal{N}_i\cup v_i}{\mathbf{X}_p} - \frac{1}{d_j + 1}\sum_{v_q\in\mathcal{N}_j\cup v_j}{\mathbf{X}_q}),
\end{split}
\end{equation}
if the network homophily is $\chi$ and further assuming that neighboring nodes strictly obey the network homophily, i.e., among $|\mathcal{N}_i\cup v_i| = d_i + 1$ neighboring nodes of the center node $v_i$, $\chi(d_i + 1)$ of them come from the same feature distribution as $v_i$ while $(1 - \chi)(d_i + 1)$ of them come from the other feature distribution as $v_j$, then symmetrically we have:
\begin{equation}
\small
    \frac{1}{d_i + 1}\sum_{v_p\in\mathcal{N}_i\cup v_i}{\mathbf{X}_p}\sim\mathcal{N}\big(\chi\boldsymbol{\mu}^{s_1} + (1 - \chi)\boldsymbol{\mu}^{s_2}, (d_i + 1)^{-1}(\chi \boldsymbol{\Sigma}^{s_1} + (1 - \chi)\boldsymbol{\Sigma}^{s_2})\big), \nonumber
\end{equation}
\begin{equation}\label{eq-dist2}
\small
    \frac{1}{d_j + 1}\sum_{v_q\in\mathcal{N}_j\cup v_j}{\mathbf{X}_q}\sim\mathcal{N}\big(\chi\boldsymbol{\mu}^{s_2} + (1 - \chi)\boldsymbol{\mu}^{s_1}, (d_j + 1)^{-1}(\chi \boldsymbol{\Sigma}^{s_2} + (1 - \chi)\boldsymbol{\Sigma}^{s_1})\big).
\end{equation}

Combining Eq.~\eqref{eq-dist2} and Eq.~\eqref{eq-dist0}, the distribution of their difference would also be a Gaussian $f_{\boldsymbol{\theta}_f}(\mathbf{X}_i) - f_{\boldsymbol{\theta}_f}(\mathbf{X}_j)\sim \mathcal{N}(\boldsymbol{\mu}, \boldsymbol{\Sigma})$, where:
\begin{equation}
\small
    \boldsymbol{\mu} = \mathbf{W}^{f, 1}\big(\chi\boldsymbol{\mu}^{s_1} + (1 - \chi)\boldsymbol{\mu}^{s_2} - \chi\boldsymbol{\mu}^{s_2} - (1-\chi)\boldsymbol{\mu}^{s_1}\big) = (2\chi - 1)\mathbf{W}^{f, 1}\Delta\boldsymbol{\mu}
\end{equation}
\begin{equation}
\small
    \boldsymbol{\Sigma} = \mathbf{W}^{f, 1}\big((d_i + 1)^{-1}(\chi \boldsymbol{\Sigma}^{s_1} + (1 - \chi)\boldsymbol{\Sigma}^{s_2}) + (d_j + 1)^{-1}(\chi \boldsymbol{\Sigma}^{s_2} + (1 - \chi)\boldsymbol{\Sigma}^{s_1})\big){\mathbf{W}^{f, 1}}^{\top}
\end{equation}

Taking the $2-$norm on the mean $\boldsymbol{\mu}$, splitting channels into sensitive ones $\mathcal{S}$ and non-sensitive ones $\mathcal{NS}$, i.e., $\{1, 2, ..., d\}=\mathcal{S}\cup\mathcal{NS}$ and expanding $\boldsymbol{\mu}$ based on the input channel, we have:
\begin{equation}\label{eq-dist3}
\small
||(2\chi - 1)\mathbf{W}^{f, 1}\Delta\boldsymbol{\mu}||_2 = (2\chi - 1)\big(\sum_{i = 1}^{d_1}(\sum_{r\in\mathcal{S}}\mathbf{W}^{f, 1}_{ir}\Delta\boldsymbol{\mu}_r + \sum_{k\in\mathcal{NS}}\mathbf{W}^{f, 1}_{ik}\Delta\boldsymbol{\mu}_k)^2\big)^{0.5},
\end{equation}
where $\mathbf{W}^{f, 1}_{ir}, \mathbf{W}^{f, 1}_{ik}$ represent the weights of the encoder from feature channel $r(k)$ to the hidden neuron $i$. Since we know that $|\mathbf{W}^{f, 1}_{ir}| \le \epsilon\mathbf{p}_r, |\mathbf{W}^{f, 1}_{ik}| \le \epsilon\mathbf{p}_k, \forall r\in\mathcal{S}, k\in\mathcal{NS}$, we substitute the upper bound here into Eq.~\eqref{eq-dist3} and finally end up with:
\begin{equation}
\footnotesize
\begin{split}
    &||\boldsymbol{\mu}||_2 = ||(2\chi - 1)\mathbf{W}^{f, 1}\Delta\boldsymbol{\mu}||_2
    \le (2\chi - 1)\big(\sum_{i = 1}^{d_1}(\sum_{r\in\mathcal{S}}\epsilon\mathbf{p}_r\Delta\boldsymbol{\mu}_r + \sum_{k\in\mathcal{NS}}\epsilon\mathbf{p}_k\Delta\boldsymbol{\mu}_k)^2\big)^{0.5}.
    \nonumber
\end{split}
\end{equation}
\vskip -2ex
\end{proof}
\vspace{-1ex}

The left side of Eq.~\eqref{eq-bound} is the difference of representations between two sensitive groups and if it is large, i.e., $||\boldsymbol{\mu}||_2$ is very large, then the predictions between these two groups would also be very different, which reflects more discrimination in terms of the group fairness. Additionally, Theorem~\ref{thm-cffairness} indicates that the upper bound of the group fairness between two sensitive groups depends on the network homophily $\chi$, the initial feature difference $\Delta\boldsymbol{\mu}$ and the masking probability $\mathbf{p}$. As the network homophily $\chi$ decreases, more neighboring nodes come from the other sensitive group and aggregating information of these neighborhoods would smooth node representations between different sensitive groups and reduce the bias. To the best of our knowledge, this is the first work relating the fairness with the network homophily. Furthermore, Eq.~\eqref{eq-bound} proves that clamping weights of the encoder $\mathbf{W}^{f, 1}$ upper bounds the group fairness.

\subsection{Training Algorithm}
Here we present a holistic algorithm of the proposed FairVGNN. In comparison to vanilla adversarial training, additional computational requirements of FairVGNN come from generating $K$ different masks. However, since within each training epoch we can pre-compute the masks as Step 4 before adversarial training and the total number of views $K$ becomes constant compared with the whole time used for adversarial training as Step 6-14, the time complexity is still linear proportional to the size of the whole graph, i.e., $O(|\mathcal{V}| + |\mathcal{E}|)$. The total model complexity includes parameters of the feature masker $O(2d)$, the discriminator/classifier $O(2d^L)$ and the encoder $O(d\prod_{l = 1}^L{d^l})$, which boils down to $O(\max_{i\in\{0, 1, ..., L\}}(d^i)^L)$ and hence the same as any other $L$-layer GNN backbones.

\vspace{-1ex}
\setlength{\textfloatsep}{4pt}
\begin{algorithm}[tbp!]
 \DontPrintSemicolon
 \footnotesize
 \KwIn{an attributed graph $G = (\mathcal{V}, \mathcal{E}, \mathbf{X}, \mathbf{A}, \mathbf{Y})$, Classifier $c_{\bm{\Theta}_c}$, Encoder $f_{\bm{\Theta}_f}$, Generator $g_{\bm{\Theta}_g}$, Discriminator $d_{\bm{\Theta}_d}$, $K$}
 
 \KwOut{Learned fairness attribute $\widetilde{\mathbf{X}}$ and Predictions $\hat{\mathbf{Y}}$}

 
 \While{not converged}{
    $\boldsymbol{\pi} \leftarrow \mathbf{W}^{g_{\bm{\Theta}_g}}$
    
    \For{$k\leftarrow 1$ \KwTo $K$}
    {
        $\mathbf{m}^k \sim \text{Gumbel-softmax}(\boldsymbol{\pi})$ , $\widetilde{\mathbf{X}}^k \leftarrow \mathbf{X}\odot\mathbf{m}^k$, \tcp*{Section~\ref{sec-g}}

        $\widetilde{\mathbf{H}}_i^{L, k} \leftarrow f_{\bm{\Theta}_f}(\widetilde{\mathbf{X}}^k, \mathbf{A})$,  $\widehat{\mathbf{H}}_i^{L, k} \leftarrow sg(\widetilde{\mathbf{H}}_i^{L, k})$    \tcp*{Section~\ref{sec-e}\footnotemark}
        
    }
    
    \For{epoch $\leftarrow 1$ \KwTo $epoch_d$}
    {
        $\mathcal{L}_{\text{d}} \leftarrow \sum\limits_{k = 1}^{K}\sum\limits_{v_i \in \mathcal{V}}[\mathbf{S}_i\log(d_{\bm{\Theta}_d}(\widehat{\mathbf{H}}^{L, k}_i)) + (1 - \mathbf{S}_i)\log(1 - d_{\bm{\Theta}_d}(\widehat{\mathbf{H}}^{L, k}_i))]$
        
        $\bm{\Theta_d} \leftarrow \bm{\Theta_d}+
        \nabla_{\bm{\Theta}_d} \mathcal{L}_d$, $\bm{\Theta_f} \leftarrow \bm{\Theta_f}+
        \nabla_{\bm{\Theta}_f}\mathcal{L}_f$ \tcp*{Section~\ref{sec-cd}}
    }
    
    \For{epoch $\leftarrow 1$ \KwTo $epoch_c$}
    {
        $\mathcal{L}_{\text{c}} \leftarrow \sum\limits_{k=1}^K\sum\limits_{v_i\in\mathcal{V}}[\mathbf{Y}_i\log(c_{\bm{\Theta}_c}(\widetilde{\mathbf{H}}^{L, k}_i)) + (1 - \mathbf{Y}_i)\log(1 - c_{\bm{\Theta}_c}(\widetilde{\mathbf{H}}^{L, k}_i))]$
        
        $\bm{\Theta_c} \leftarrow \bm{\Theta_c}-
        \nabla_{\bm{\Theta}_c}\mathcal{L}_c$, $\bm{\Theta_f} \leftarrow \bm{\Theta_f}-
        \nabla_{\bm{\Theta}_f}\mathcal{L}_f$ \tcp*{Section~\ref{sec-cd}}
    }
    
    \For{epoch $\leftarrow 1$ \KwTo $epoch_g$}
    {
        $\mathcal{L}_{\text{g}}^k \leftarrow \sum\limits_{k=1}^K\sum\limits_{v_i\in\mathcal{V}}||d_{\bm{\Theta}_d}(\widetilde{\mathbf{H}}^{L, k}_i) - 0.5||_2^2,$
    
        $\bm{\Theta_g} \leftarrow \bm{\Theta_g}-
        \nabla_{\bm{\Theta}_g}\mathcal{L}_g$, $\bm{\Theta_f} \leftarrow \bm{\Theta_f}-
        \nabla_{\bm{\Theta}_f}\mathcal{L}_f$ \tcp*{Section~\ref{sec-advtraining}}
    }
    
    $\bm{\Theta_f} \leftarrow \text{Clamp}(\bm{\Theta_f}, \sum\limits_{k = 1}^K\mathbf{m}^k)$ \tcp*{Section~\ref{sec-wc}}
    
}

$\widetilde{\mathbf{X}} = \sum\limits_{k=1}^K\widetilde{\mathbf{X}}^k, ~~\hat{\mathbf{Y}} = c_{\bm{\Theta}_c}(f_{\bm{\Theta}_f}(\widetilde{\mathbf{X}}, \mathbf{A}))$~~~~

\KwRet{$\widetilde{\mathbf{X}}, \widehat{\mathbf{Y}}$}

\caption{\small The algorithm of FairVGNN}
\label{alg-fairvgnn}
\end{algorithm}
\footnotetext{$sg$: stopgrad prevents gradients from being
back-propagated.}

%% file: experiments.tex
\section{Experiments}\label{sec-experiment}
In this section, we conduct extensive experiments to evaluate the effectiveness of FairVGNN.
\subsection{Experimental Settings}\label{sec-experimentsetting}

\subsubsection{Datasets}\label{sec-dataset}
We validate the proposed approach on three benchmark datasets~\cite{nifty, EDITS} with their statistics shown in Table~\ref{tab-dataset}.
\begin{table}[htbp!]
\vskip -2ex
\small
\setlength{\extrarowheight}{.095pt}
\setlength\tabcolsep{3pt}
\caption{Basic dataset statistics.}
\centering
\vspace{-4ex}
\begin{tabular}{lccc}
 \Xhline{2\arrayrulewidth}
\textbf{Dataset} & \textbf{German}& \textbf{Credit} & \textbf{Bail}\\
 \Xhline{1.5\arrayrulewidth}
\#Nodes & 1000& 30,000 & 18,876 \\
\#Edges & 22,242 & 1,436,858 & 321,308\\
\#Features & 27 & 13 & 18\\
Sens. & Gender & Age & Race\\
Label & Good/bad Credit & Default/no default Payment & Bail/no bail\\

\Xhline{2\arrayrulewidth}
\end{tabular}
\label{tab-dataset}
\vskip -1ex
\end{table}

\begin{table*}[t!]
\tiny
\setlength{\extrarowheight}{.095pt}
\setlength\tabcolsep{3pt}
\centering
\caption{Model utility and bias of node classification. We compare the proposed FairVGNN (i.e., FairV) against state-of-the-art baselines NIFTY, EDITS, and FairGNN (i.e., Fair) when equiped with various GNN backbones (i.e., GCN, GIN, and SAGE). The best and runner-up results are colored in \textcolor{red}{red} and \textcolor{blue}{blue}. $\uparrow$ represents the larger, the better while $\downarrow$ represents the opposite.}
\vskip -3.5ex
\begin{tabular}{l|l|ccccc|ccccc|ccccc|c}
\hline
 \multirow{2}{*}{\textbf{Encoder}}& \multirow{2}{*}{\textbf{Method}} & \multicolumn{5}{c|}{\textbf{German}} & \multicolumn{5}{c|}{\textbf{Credit}} & \multicolumn{5}{c|}{\textbf{Bail}} & \multirow{2}{*}{
 \begin{tabular}{@{}c@{}} \textbf{Avg.} \\ \textbf{(Rank)}\end{tabular}
  } \\
 &  & AUC ($\uparrow$) & F1 ($\uparrow$) & ACC ($\uparrow$) & $\Delta_{\text{sp}}$ ($\downarrow$) & $\Delta_{\text{eo}}$ ($\downarrow$) & AUC ($\uparrow$) & F1 ($\uparrow$) & ACC ($\uparrow$) & $\Delta_{\text{sp}}$ ($\downarrow$) & $\Delta_{\text{eo}}$ ($\downarrow$) & AUC ($\uparrow$) & F1 ($\uparrow$) & ACC ($\uparrow$) & $\Delta_{\text{sp}}$ ($\downarrow$) & $\Delta_{\text{eo}}$ ($\downarrow$) \\
 
 \hline
 
\multirow{5}{*}{\textbf{GCN}} & Vanilla & \textcolor{blue}{74.11$\pm$0.37} & 82.46$\pm$0.89 & \textcolor{blue}{73.44$\pm$1.09} & 35.17$\pm$7.27 & 25.17$\pm$5.89 & 73.87$\pm$0.02 & 81.92$\pm$0.02 & 73.67$\pm$0.03 & 12.86$\pm$0.09 & 10.63$\pm$0.13 & 87.08$\pm$0.35 & 79.02$\pm$0.74 & 84.56$\pm$0.68 & 7.35$\pm$0.72 & 4.96$\pm$0.62 & 9.17\\

 & NIFTY & 68.78$\pm$2.69 & 81.40$\pm$0.54 & 69.92$\pm$1.14 & 5.73$\pm$5.25 & 5.08$\pm$4.29 & 71.96$\pm$0.19 & 81.72$\pm$0.05 & 73.45$\pm$0.06 & 11.68$\pm$0.07 & 9.39$\pm$0.07 & 78.20$\pm$2.78 & 64.76$\pm$3.91 & 74.19$\pm$2.57 & 2.44$\pm$1.29 & 1.72$\pm$1.08 & 9.69\\
 
& EDITS & 69.41$\pm$2.33 & 81.55$\pm$0.59 & 71.60$\pm$0.89 & 4.05$\pm$4.48 & 3.89$\pm$4.23 & 73.01$\pm$0.11 & 81.81$\pm$0.28 & 73.51$\pm$0.30 & 10.90$\pm$1.22 & 8.75$\pm$1.21 & 86.44$\pm$2.17 & 75.58$\pm$3.77 & 84.49$\pm$2.27 & 6.64$\pm$0.39 & 7.51$\pm$1.20 & 9.89\\
  
  
 & FairGNN & 67.35$\pm$2.13 & 82.01$\pm$0.26 & 69.68$\pm$0.30 & 3.49$\pm$2.15 & 3.40$\pm$2.15 & 71.95$\pm$1.43 & 81.84$\pm$1.19 & 73.41$\pm$1.24 & 12.64$\pm$2.11 & 10.41$\pm$2.03 & 87.36$\pm$0.90 & 77.50$\pm$1.69 & 82.94$\pm$1.67 & 6.90$\pm$0.17 & 4.65$\pm$0.14 & 9.17\\
 
 & FairVGNN  & 72.41$\pm$2.10& 82.14$\pm$0.42 & 70.16$\pm$0.86 & 1.71$\pm$1.68 & \textcolor{blue}{0.88$\pm$0.58} & 71.34$\pm$0.41 & 87.08$\pm$0.74 & 78.04$\pm$0.33 & 5.02$\pm$5.22 & 3.60$\pm$4.31 & 85.68$\pm$0.37 & 79.11$\pm$0.33 & 84.73$\pm$0.46 & 6.53$\pm$0.67 & 4.95$\pm$1.22 & 5.67\\
 
 \hline
 
\multirow{5}{*}{\textbf{GIN}} & Vanilla & 72.71$\pm$1.44 & \textcolor{blue}{82.78$\pm$0.50} & \textcolor{red}{73.84$\pm$0.54} & 13.56$\pm$5.23 & 9.47$\pm$4.49 & 74.36$\pm$0.21 & 82.28$\pm$0.64 & 74.02$\pm$0.73 & 14.48$\pm$2.44 & 12.35$\pm$2.86 & 86.14$\pm$0.25 & 76.49$\pm$0.57 & 81.70$\pm$0.67 & 8.55$\pm$1.61 & 6.99$\pm$1.51 & 9.56\\

 & NIFTY & 67.61$\pm$4.88 & 80.46$\pm$3.06 & 69.92$\pm$3.64 & 5.26$\pm$3.24 & 5.34$\pm$5.67 & 70.90$\pm$0.24 & 84.05$\pm$0.82 & 75.59$\pm$0.66 & 7.09$\pm$4.62 & 6.22$\pm$3.26 & 82.33$\pm$4.61 & 70.64$\pm$6.73 & 74.46$\pm$9.98 & 5.57$\pm$1.11 & 3.41$\pm$1.43 & 8.56\\

 & EDITS & 69.35$\pm$1.64 & 82.80$\pm$0.22 & 72.08$\pm$0.66 & 0.86$\pm$0.76 & 1.72$\pm$1.14 & 72.35$\pm$1.11 & 82.47$\pm$0.85 & 74.07$\pm$0.98 & 14.11$\pm$14.45 & 15.40$\pm$15.76 & 80.19$\pm$4.62 & 68.07$\pm$5.30 & 73.74$\pm$5.12 & 6.71$\pm$2.35 & 5.98$\pm$3.66 & 11.36\\
 
 & FairGNN & 72.95$\pm$0.82 & \textcolor{red}{83.16$\pm$0.56} & 72.24$\pm$1.44 & 6.88$\pm$4.42 & 2.06$\pm$1.46 & 68.66$\pm$4.48 & 79.47$\pm$5.29 & 70.33$\pm$5.50 & \textcolor{blue}{4.67$\pm$3.06} & 3.94$\pm$1.49 & 86.14$\pm$0.89& 73.67$\pm$1.17& 77.90$\pm$2.21& 6.33$\pm$1.49& 4.74$\pm$1.64 & 7.64\\
 
 & FairVGNN & 71.65$\pm$1.90 & 82.40$\pm$0.14 & 70.16$\pm$ 0.32 & \textcolor{red}{0.43$\pm$0.54} & \textcolor{red}{0.34$\pm$0.41} & 71.36$\pm$0.72 & \textcolor{blue}{87.44$\pm$0.23} & \textcolor{blue}{78.18$\pm$0.20} & \textcolor{red}{2.85$\pm$2.01} & \textcolor{red}{1.72$\pm$1.80} & 83.22$\pm$1.60 & 76.36$\pm$2.20 & 83.86$\pm$1.57 & 5.67$\pm$0.76 & 5.77$\pm$1.26 & \textcolor{blue}{5.44}\\
 
 \hline
 
\multirow{5}{*}{\textbf{SAGE}} & Vanilla & \textcolor{red}{75.74$\pm$0.69} & 81.25$\pm$1.72 & 72.24$\pm$1.61 & 24.30$\pm$6.93 & 15.55$\pm$7.59 & \textcolor{red}{74.58$\pm$1.31} & 83.38$\pm$0.77 & 75.28$\pm$0.83 & 15.65$\pm$1.30 & 13.34$\pm$1.34 & 90.71$\pm$0.69 & 80.99$\pm$0.55 & 86.72$\pm$0.48 & 2.16$\pm$1.53 & \textcolor{red}{0.84$\pm$0.55} & 7.31\\

 & NIFTY & 72.05$\pm$2.15 & 79.20$\pm$1.19 & 69.60$\pm$1.50 & 7.74$\pm$7.80 & 5.17$\pm$2.38 & 72.89$\pm$0.44 & 82.60$\pm$1.25 & 74.39$\pm$1.35 & 10.65$\pm$1.65 & 8.10$\pm$1.91 & \textcolor{red}{92.04$\pm$0.89} & 77.81$\pm$6.03 & 84.11$\pm$5.49 & 5.74$\pm$0.38 & 4.07$\pm$1.28 & 8.06 \\
 
 & EDITS & 69.76$\pm$5.46 & 81.04$\pm$1.09 & 71.68$\pm$1.25 & 8.42$\pm$7.35 & 5.69$\pm$2.16 & 75.04$\pm$0.12 & 82.41$\pm$0.52 & 74.13$\pm$0.59 & 11.34$\pm$6.36 & 9.38$\pm$5.39 & 89.07$\pm$2.26 & 77.83$\pm$3.79 & 84.42$\pm$2.87 & 3.74$\pm$3.54 & 4.46$\pm$3.50 & 11.36\\
 
 & FairGNN & 65.85$\pm$9.49 & 82.29$\pm$0.32 & 70.64$\pm$0.74 & 7.65$\pm$8.07 & 4.18$\pm$4.86 & 70.82$\pm$0.74  & 83.97$\pm$2.00 & 75.29$\pm$1.62 & 6.17$\pm$5.57 & 5.06$\pm$4.46 & 91.53$\pm$0.38 & \textcolor{blue}{82.55$\pm$0.98} & \textcolor{blue}{87.68$\pm$0.73} & \textcolor{blue}{1.94$\pm$0.82} & 1.72$\pm$0.70 & 5.83\\
 
 & FairVGNN & 73.84$\pm$0.52 & 81.91$\pm$0.63 & 70.00$\pm$0.25 & \textcolor{blue}{1.36$\pm$1.90} & 1.22$\pm$1.49 & \textcolor{blue}{74.05$\pm$0.20} & \textcolor{red}{87.84$\pm$0.32} & \textcolor{red}{79.94$\pm$0.30} & 4.94$\pm$1.10 & \textcolor{blue}{2.39$\pm$0.71} & \textcolor{blue}{91.56$\pm$1.71} & \textcolor{red}{83.58$\pm$1.88} & \textcolor{red}{88.41$\pm$1.29} & \textcolor{red}{1.14$\pm$0.67} & \textcolor{blue}{1.69$\pm$1.13} & \textcolor{red}{2.92}\\
 \hline
\end{tabular}
\label{tab-main}
\vskip -3ex
\end{table*}


\subsubsection{Baselines}
Several state-of-the-art fair node representation learning models are compared with our proposed FairVGNN. We divide them into 
two categories:
\begin{inparaenum}
    \item \textbf{Augmentation-based}: this type of methods alleviates discrimination via graph augmentation, where sensitive-related information is removed by modifying the graph topology or node features. NIFTY~\cite{nifty} simultaneously achieves the Counterfactual Fairness and the stability by contrastive learning. EDITS~\cite{EDITS} approximates the inputs' discrimination via Wasserstein distance and directly minimizes it between sensitive and non-sensitive groups by pruning the graph topology and node features.
    \item \textbf{Adversarial-based}:  The adversarial-based methods enforce the fairness of node representations by alternatively training the encoder to fool the discriminator and the discriminator to predict the sensitive attributes. FairGNN~\cite{dai2021say} deploys an extra sensitive feature estimator to increase the amount of sensitive information
\end{inparaenum}
Since different GNN-backbones may cause different levels of sensitive attribute leakage, 
we consider to equip each of the above three bias-alleviating methods with three GNN-backbones: GCN~\cite{GCN}, GIN~\cite{GIN}, GraphSAGE~\cite{Graphsage}, e.g., GCN-NIFTY represents the GCN encoder with NIFTY. 
\vspace{-1.5ex}
\subsubsection{Setup}
Our proposed FairVGNN is implemented using PyTorch-Geometric~\cite{pytorch}. For EDITS\footnote{\url{https://github.com/yushundong/edits}},
NIFTY\footnote{\url{https://github.com/chirag126/nifty}} and FairGNN\footnote{\url{https://github.com/EnyanDai/FairGNN}}, we use the original code from the authors' GitHub repository. We aim to provide a rigorous and fair comparison between different models on each dataset by tuning hyperparameters for all models individually and detailed hyperparamter configuration of each baseline is in Appendix~\ref{app-hyper}. Following~\cite{nifty} and~\cite{EDITS}, we use 1-layer GCN, GIN convolution and 2-layer GraphSAGE convolution respectively as our encoder $f_{\bm{\Theta}_f}$, and use 1 linear layer as our classifier $c_{\bm{\Theta}_c}$ and discriminator $d_{\bm{\Theta}_d}$. The detailed GNN architecture is described in Appendix~\ref{app-architecture}. We fix the number of hidden unit of the encoder $f_{\bm{\Theta}_f}$ as 16, the dropout rate as 0.5, the number of generated fair feature views during each training epoch $K=10$. The learning rates and the training epochs of the generator $g_{\bm{\Theta}_g}$, the discriminator $d_{\bm{\Theta}_d}$, the classifier $c_{\bm{\Theta}_c}$ and the encoder $f_{\bm{\Theta}_f}$ are searched from $\{0.001, 0.01\}$ and $\{5, 10\}$, the prefix cutting threshold $\epsilon$ in Eq.~\eqref{eq-wc} is searched from $\{0.01, 0.1, 1\}$, the whole training epochs as $200, 300, 400$, and $\alpha \in \{0, 0.5, 1\}$. We use the default data splitting following~\cite{nifty, EDITS} and experimental results are averaged over five repeated executions with five different seeds to remove any potential initialization bias.


\vspace{-2ex}
\subsection{Node Classification}
\vspace{-0.5ex}
\subsubsection{Performance comparison}
The model utility and fairness of each baseline is shown in Table~\ref{tab-main}. We observe that our FairVGNN consistently performs the best compared with other bias-alleviating methods in terms of the average rank for all datasets and across all evaluation metrics, which indicates the superiority of our model in achieving better trade-off between model utility and fairness. 
Since no fairness regularization is imposed on GNN encoders equipped with vanilla methods, they generally achieve better model utility. However for this reason, sensitive-related information is also completely free to be encoded in the learned node representations and hence causes higher bias. To alleviate such discrimination, all other methods propose different regularizations to constrain sensitive-related information in learned node representations, which also remove some task-related information and hence sacrifice model utility as expected in Table~\ref{tab-main}. However, we do observe that our model can yield lower biased predictions
with less utility sacrifice, which is mainly ascribed to two reasons:
\begin{inparaenum}
    \item We generate different fair feature views by randomly sampling masks from learned Gumbel-Softmax distribution and make predictions. This can be regarded as a data augmentation technique by adding noise to node features, which decreases the population risk and enhances the model generalibility~\cite{shorten2019survey} by creating novel mapping from augmented training points to the label space.
    \item The weight clamping module clamps weights of encoder based on feature correlations to the sensitive feature channel, which adaptively remove/keep the sensitive/task-relevant information.
\end{inparaenum}

\begin{table*}[t!]
\tiny
\setlength{\extrarowheight}{.095pt}
\setlength\tabcolsep{3pt}
\centering
\caption{Model utility and bias of node classification of different variants of FairVGNN. The best and runner-up results are colored in \textcolor{red}{red} and \textcolor{blue}{blue}. $\uparrow$ represents the larger, the better while $\downarrow$ represents the opposite.}
\vskip -3.5ex
\begin{tabular}{l|l|ccccc|ccccc|cccccc}
\hline
\multirow{2}{*}{\textbf{Encoder}} & \multirow{2}{*}{\textbf{Model Variants}} & \multicolumn{5}{c|}{\textbf{German}} & \multicolumn{5}{c|}{\textbf{Credit}} & \multicolumn{5}{c}{\textbf{Bail}} \\
 &  & AUC ($\uparrow$) & F1 ($\uparrow$) & ACC ($\uparrow$) & $\Delta_{\text{sp}}$ ($\downarrow$) & $\Delta_{\text{eo}}$ ($\downarrow$) & AUC ($\uparrow$) & F1 ($\uparrow$) & ACC ($\uparrow$) & $\Delta_{\text{sp}}$ ($\downarrow$) & $\Delta_{\text{eo}}$ ($\downarrow$) & AUC ($\uparrow$) & F1 ($\uparrow$) & ACC ($\uparrow$) & $\Delta_{\text{sp}}$ ($\downarrow$) & $\Delta_{\text{eo}}$ ($\downarrow$) \\
 \hline
\multirow{4}{*}{\textbf{GCN}} & FairV & 72.69$\pm$ 1.67 & 81.86$\pm$ 0.49 & 69.84$\pm$0.41 & 0.77$\pm$ 0.39 & 0.46$\pm$ 0.34 & 71.34$\pm$0.41 & 87.08$\pm$0.74 & 78.04$\pm$0.33 & 5.02$\pm$5.22 & 3.60$\pm$4.31 & 85.68$\pm$0.37 & 79.11$\pm$0.33 & 84.73$\pm$0.46 & 6.53$\pm$0.67 & 4.95$\pm$1.22 \\
 & FairV w/o fm & 73.63$\pm$ 1.14 & 82.28$\pm$0.28 & 70.88$\pm$1.09 & 5.56$\pm$3.89 & 4.41$\pm$3.59 & 72.51$\pm$0.32 & 86.15$\pm$2.18 & 77.83$\pm$2.15 & 6.94$\pm$2.86 & 4.64$\pm$2.73 & 86.98$\pm$0.32 & 78.08$\pm$0.53 & 84.59$\pm$0.29 & 7.24$\pm$0.26 & 5.75$\pm$0.68 \\
 & FairV w/o wc & 72.08$\pm$ 1.83 & 82.72$\pm$ 0.50 & 71.04$\pm$ 1.23 & 3.19$\pm$ 3.51 & 0.59$\pm$ 1.12 & 71.80$\pm$0.47 & 87.27$\pm$0.47 & 78.47$\pm$0.34 & 9.05$\pm$4.55 & 5.94$\pm$3.61 & 85.93$\pm$0.38 & 79.22$\pm$0.29 & 85.38$\pm$0.25 & 6.61$\pm$0.48 & 5.82$\pm$0.66 \\
 & FairV w/o fm\&wc & 74.97$\pm$0.94 & 82.30$\pm$0.67 & 70.8$\pm$0.88 & 7.74$\pm$5.05 & 4.56$\pm$4.15 & 73.09$\pm$0.41 & 84.48$\pm$2.14 & 76.40$\pm$2.29 & 11.91$\pm$2.34 & 9.27$\pm$1.98 & 86.44$\pm$0.16 & 78.75$\pm$0.27 & 84.41$\pm$0.28 & 8.32$\pm$0.60 & 6.34$\pm$0.32 \\
 \hline
\multirow{4}{*}{\textbf{GIN}} & FairV & 71.65$\pm$1.90 & 82.40$\pm$0.14 & 70.16$\pm$0.32 & \textcolor{red}{0.43$\pm$0.54} & \textcolor{red}{0.34$\pm$0.41} & 71.36$\pm$0.72 & 87.44$\pm$0.23 & 78.18$\pm$0.20 & \textcolor{red}{2.85$\pm$2.01} & \textcolor{blue}{1.72$\pm$1.80} & 83.22$\pm$1.60 & 76.36$\pm$2.20 & 83.86$\pm$1.57 & 5.67$\pm$0.76 & 5.77$\pm$1.26 \\
 & FairV w/o fm & 73.76$\pm$0.77 & \textcolor{blue}{83.06$\pm$0.67} & \textcolor{blue}{71.68$\pm$1.63} & 2.76$\pm$2.64 & \textcolor{blue}{0.57$\pm$0.47} & 71.15$\pm$0.63 & 87.09$\pm$0.7 & 78.29$\pm$0.53 & 3.36$\pm$2.34 & 1.86$\pm$1.19 & 85.12$\pm$0.54 & 77.06$\pm$0.83 & 83.13$\pm$1.19 & 6.80$\pm$0.28 & 5.97$\pm$0.64 \\
 & FairV w/o wc & 72.65$\pm$1.65 & 82.70$\pm$0.30 & 71.20$\pm$1.01 & 3.44$\pm$3.19 & 0.97$\pm$0.9 & 71.13$\pm$0.59 & \textcolor{blue}{87.96$\pm$0.25} & \textcolor{blue}{80.04$\pm$0.22} & \textcolor{blue}{3.16$\pm$1.28} & \textcolor{red}{1.47$\pm$0.72} & 85.09$\pm$2.36 & 79.07$\pm$2.70 & \textcolor{blue}{85.85$\pm$2.13} & 5.24$\pm$1.41 & 4.33$\pm$2.05 \\
 & FairV w/o fm\&wc & 73.41$\pm$1.17 & \textcolor{red}{83.20$\pm$0.44} & \textcolor{red}{72.40$\pm$1.29} & 5.70$\pm$4.57 & 1.01$\pm$1 & 72.73$\pm$0.32 & 86.10$\pm$0.59 & 77.90$\pm$0.63 & 6.66$\pm$1.10 & 3.97$\pm$0.41 & 86.32$\pm$1.60 & 79.28$\pm$1.39 & \textcolor{red}{86.02$\pm$0.40} & 7.48$\pm$0.71 & 7.43$\pm$2.38 \\
 \hline
\multirow{4}{*}{\textbf{SAGE}} & FairV & 73.84$\pm$0.52 & 81.91$\pm$0.63 & 70.00$\pm$0.25 & \textcolor{blue}{1.36$\pm$1.90} & 1.22$\pm$1.49 & 74.05$\pm$0.20 & 87.84$\pm$0.32 & 79.94$\pm$0.30 & 4.94$\pm$1.10 & 2.39$\pm$0.71 & 91.56$\pm$1.71 & 83.58$\pm$1.88 & 88.41$\pm$1.29 & \textcolor{red}{1.14$\pm$0.67} & 1.69$\pm$1.13 \\
 & FairV w/o fm & \textcolor{blue}{73.98$\pm$1.40} & 81.36$\pm$1.45 & 70.00$\pm$1.50 & 3.67$\pm$2.80 & 1.55$\pm$2.01 & 73.58$\pm$0.68 & 83.18$\pm$2.32 & 74.97$\pm$2.49 & 7.23$\pm$3.91 & 5.05$\pm$3.17 & 91.96$\pm$0.57 & 84.04$\pm$1.01 & \textcolor{blue}{88.69$\pm$0.79} & \textcolor{blue}{1.51$\pm$1.17} & \textcolor{blue}{1.59$\pm$0.35} \\
 & FairV w/o wc & 73.93$\pm$2.16 & 82.02$\pm$0.72 & 70.16$\pm$1.25 & 2.80$\pm$2.79 & 0.90$\pm$1.06 & \textcolor{blue}{74.05$\pm$0.42} & \textcolor{red}{88.10$\pm$0.30} & \textcolor{red}{80.16$\pm$0.19} & 5.09$\pm$1.30 & 2.67$\pm$0.92 & \textcolor{blue}{92.01$\pm$0.74} & 84.64$\pm$0.91 & \textcolor{red}{89.24$\pm$0.58} & 2.99$\pm$0.94 & \textcolor{red}{1.07$\pm$1.19} \\
 & FairV w/o fm\&wc & 73.87$\pm$1.62 & 80.09$\pm$1.73 & 70.08$\pm$1.17 & 6.18$\pm$1.31 & 4.68$\pm$2.38 & \textcolor{red}{74.57$\pm$0.14} & 81.91$\pm$0.92 & 73.61$\pm$1.02 & 7.27$\pm$3.22 & 5.03$\pm$3.01 & \textcolor{red}{92.05$\pm$0.89} & 83.40$\pm$1.79 & 88.44$\pm$1.02 & 3.51$\pm$0.87 & 2.05$\pm$1.19\\
 \hline
\end{tabular}
\label{tab-ablation}
\vskip -3.5ex
\end{table*}

\vspace{-1.5ex}
\subsubsection{Ablation study}
Next we conduct ablation study to fully understand the effect of each component of FairVGNN on alleviating discrimination. Concretely, we denote \textbf{FairV w/o fm} as removing the module of generating fair feature views, \textbf{FairV w/o ad wc} as removing the module of adaptive weight clamping, and \textbf{FairV w/o fm\&ad wc} as removing both of these two modules. Since computing thresholds in adaptive weight clamping needs the probability of feature masking from fair feature view generation in Eq.~\eqref{eq-wc}, we instead directly take the prefix value $\epsilon$ without $\mathbf{p}_i$ as our cutting threshold in \textsl{FairV w/o fm}. The utility and bias of these variants are presented in Table~\ref{tab-ablation}. We observe that \textsl{FairV w/o fm} and \textsl{FairV w/o ad wc} perform worse than \textsl{FairV}, which validates the effectiveness of different components in \textsl{FairV} for learning fair node representations. Furthermore, the  worse performance of \textsl{FairV w/o fm\&ad wc} than \textsl{FairV w/o fm} and \textsl{FairV w/o wc} indicates the proposed two modules alleviate discrimination from two different aspects and their effects could be accumulated together. In most cases, \textsl{FairV w/o fm} achieves more bias than \textsl{FairV w/o ad wc}. This is because the original clamping threshold of sensitive feature channels $\epsilon*\mathbf{p}_i$ would be replaced by a higher threshold $\epsilon$, which allows more sensitive information leakage to predictions.


\vspace{-2.25ex}
\subsection{Further Probe}

\vspace{-1ex}
\subsubsection{Does adversarial training work?}
\begin{figure}[t]
    \vskip -1ex
     \centering
     \hspace{-1ex}
     \includegraphics[width=0.45\textwidth]{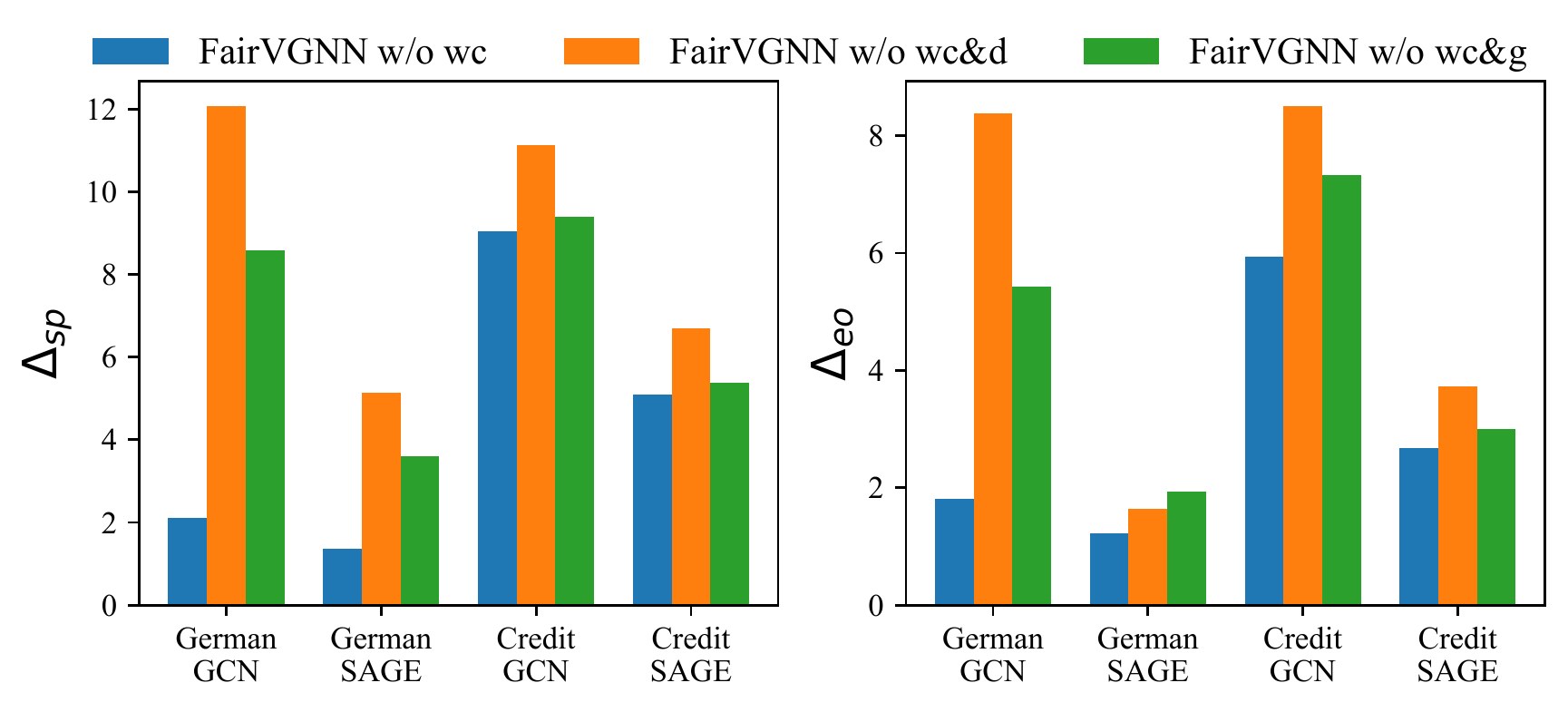}
     \vskip -2.75ex
      \caption{\hspace{-0.5ex} Model bias without the discriminator/generator.}
     \label{fig-adv_ablation}
    \vspace{-0.5ex}
\end{figure}

We first remove the weight clamping to solely study the effect of adversarial training, and then remove the discriminator/generator respectively by setting their corresponding training epochs to be 0 and denote the corresponding models as \textbf{FairVGNN w/o wc\&d} and \textbf{FairVGNN w/o wc\&g}. We re-conduct the node classification with five different initializations following the previous setting and report the average bias in Figure~\ref{fig-adv_ablation}. We can clearly see that after removing discriminator or generator, the model bias becomes even higher in both  situations, which indicates the importance of the competition between the discriminator and the generator in improving the discriminative power of discriminator to recognize sensitive features and the generating power of generator to generate fair feature views. Moreover, since the discriminator in \textsl{FairVGNN w/o wc\&g} can still recognize the sensitive features and then guide the encoder to extract less sensitive-related information, the bias of \textsl{FairVGNN w/o wc\&g} is lower than \textsl{FairVGNN w/o wc\&d} in most cases.

\vspace{-1.5ex}
\subsubsection{Does adaptive weight clamping work?}
To demonstrate the advantages of the proposed adaptive weight clamping, here we compare it with the non-adaptive weight clamping and spectral normalization, which is another technique of regularizing weight matrix to enhance the model robustness and counterfactual fairness~\cite{nifty}. The prefix cutting thresholds in both the adaptive and non-adaptive weight clamping are set to be the same as the best ones tuned in Table~\ref{tab-main} for SAGE/GCN/GIN to ensure the fair comparison. As shown in Table~\ref{tab-fairnesswc}, we can see that except for GIN, the adaptive weight clamping always achieves lower bias while not hurting so much model utility. This is because for sensitive-related feature channels, multiplying masking probability by the prefix threshold would even lower the threshold and prevent more sensitive information from leaking to prediction through the encoder.
We also investigate the influence of prefix cutting threshold $\epsilon$ in Eq.~\eqref{eq-wc} on the model bias/utility. Higher $\epsilon$ indicates less weight clamping on the encoder and more sensitive-related information is leveraged in predictions, which leads to higher bias.

\begin{table}[t]
\footnotesize
\setlength{\extrarowheight}{.11pt}
\setlength\tabcolsep{3pt}
\centering
\caption{Comparison with different weight regularization.}
\vspace{-2.5ex}
\label{tab-fairnesswc}
\begin{tabular}{l|l|ccccc}
\hline
\begin{tabular}{@{}c@{}} \textbf{Dataset} \\ \textbf{(Model)}\end{tabular} & \multirow{1}{*}{\textbf{Strategy}} & AUC ($\uparrow$) & ACC ($\uparrow$) & F1 ($\uparrow$) & $\Delta_{\text{sp}}$ ($\downarrow$) & $\Delta_{\text{eo}}$ ($\downarrow$) \\
 \hline
\multirow{3}{*}{\begin{tabular}{@{}c@{}} \textbf{German} \\ \textbf{(SAGE)}\end{tabular}} & Ad wc & \textcolor{red}{73.84+0.52} & 70.00+0.25 & 81.91+0.63 & \textcolor{red}{1.36+1.90} & \textcolor{red}{1.22+1.49} \\
 & Wc & 72.43+1.60 & \textcolor{red}{70.48+0.85} & \textcolor{red}{82.03+0.82} & 4.85+4.10 & 2.50+2.12 \\
 & Sn & 73.00+1.53 & 70.00+1.07 & 81.82+0.59 & 3.74+3.22 & 1.89+1.08 \\
 \hline
 
\multirow{3}{*}{\begin{tabular}{@{}c@{}} \textbf{Credit} \\ \textbf{(GIN)}\end{tabular}}  & Ad wc & \textcolor{red}{74.05$\pm$0.20} & \textcolor{red}{79.94$\pm$0.19} & \textcolor{red}{87.84$\pm$0.32} & 4.94$\pm$1.10 & 2.39$\pm$0.71 \\
 & Wc & 73.20$\pm$1.20 & 79.03$\pm$1.09 & 87.23$\pm$0.94 & 7.03$\pm$4.58 & 4.74$\pm$3.47 \\
 & Sn & 71.12$\pm$0.55 & 78.54$\pm$2.00 & 86.53$\pm$1.90 & \textcolor{red}{2.60$\pm$0.73} & \textcolor{red}{0.87$\pm$0.54}\\
 \hline
 
\multirow{3}{*}{\begin{tabular}{@{}c@{}} \textbf{Bail} \\ \textbf{(GCN)}\end{tabular}} & Ad wc & 85.68+0.37 & 84.73+0.46 & 79.11+0.33 & \textcolor{red}{6.53+0.67} & \textcolor{red}{4.95+1.22} \\
 & Wc & 85.97+0.45 & 85.12+0.26 & 79.08+0.28 & 6.86+0.47 & 5.85+0.83 \\
 & Sn & \textcolor{red}{86.10+0.61} & \textcolor{red}{85.69+0.42} & \textcolor{red}{79.66+0.63} & 7.53+0.17 & 6.43+0.81\\
 \hline

\end{tabular}
\begin{tablenotes}
  \scriptsize
  \item \textbf{*} \textbf{Ad wc}: adaptively clamp weights of the encoder; \textbf{Wc}: clamp weights of the encoder; and \\\hspace{2.25ex}\textbf{Sn}: spectral normalization of the encoder
\end{tablenotes}
\vskip -4ex
\end{table}

\begin{figure}[t]
     \centering
     \hspace{-3.5ex}
     \begin{subfigure}[b]{0.235\textwidth}
         \centering
         \includegraphics[width=1.07\textwidth]{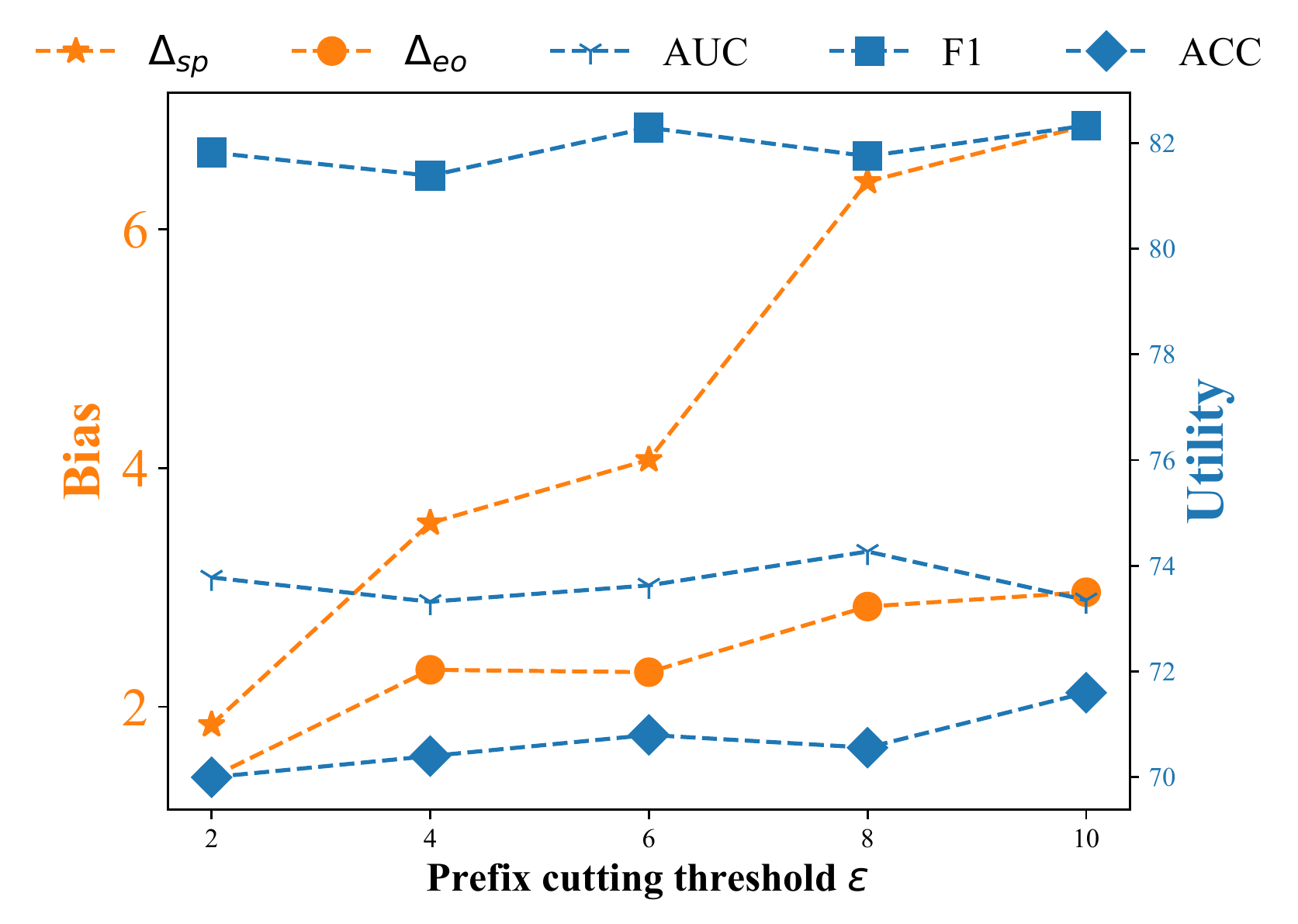}
         \vskip -0.25ex
         \caption{German}
         \label{fig-germansagewc}
     \end{subfigure}
     \hspace{.25ex}
     \begin{subfigure}[b]{0.235\textwidth}
         \centering
         \includegraphics[width=1.07\textwidth]{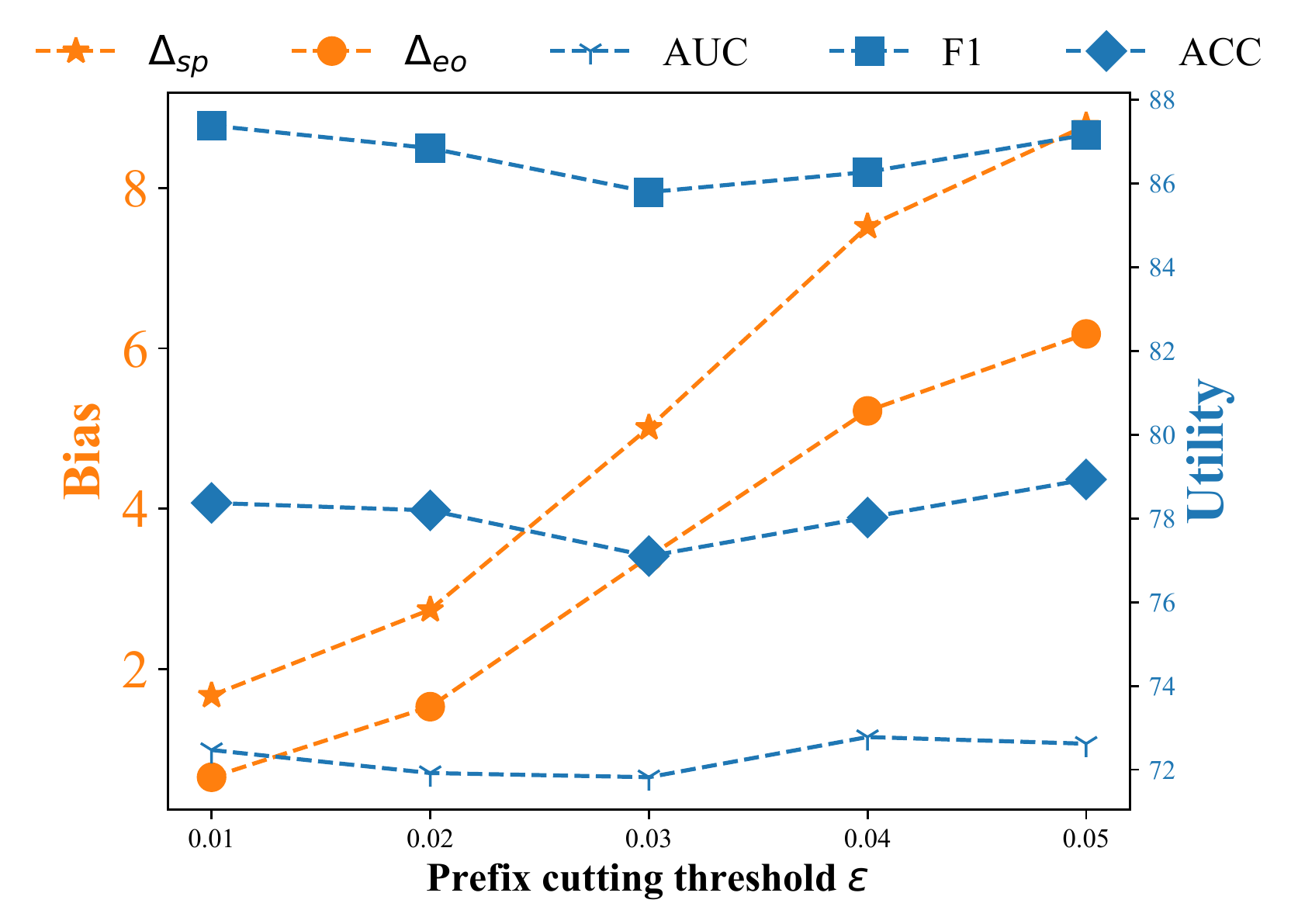}
         \vskip -0.25ex
         \caption{Credit}
         \label{fig-creditsagewc}
     \end{subfigure}
     \vskip -2.25ex
     \caption{Results of different prefix cutting threshold.}
     \label{fig-layer}
     \vskip -1ex
\end{figure}



%% file: relatedwork.tex
\vspace{-1ex}
\section{Related Work}\label{sec-relatedwork}

Most prior work on GNNs exclusively focus on optimizing the model utility while totally ignoring the bias encoded in the learned node representations, which would unavoidably cause social risks in high-stake discriminatory decisions~\cite{EDITS}. 
FairGNN~\cite{dai2021say} leverages a sensitive feature estimator to enhance the amount of the sensitive attributes, 
which greatly benefits their adversarial debiasing procedure. NIFTY~\cite{nifty} proposes a novel triplet-based objective function and a layer-wise weight normalization using the Lipschitz constant to promote counterfactual fairness and stability of the resulted node representations. EDITS~\cite{EDITS} systematically summarizes the biased node representation learning into attribute bias and structure bias, and employs the Wasserstein distance approximator to alternately debias node features and network topology. More recently, REFEREE~\cite{REFEREE} was proposed to provide structural explanations of bias in GNNs. Different from previous work, we study a novel problem that feature propagation could cause correlation variation and sensitive leakage to innocuous features, and our proposed framework FairVGNN expects to learn which feature channels should be masked to alleviate discrimination considering the effect of correlation variation. Recently, others have also explored this concept of varying correlation during feature propagation towards developing deeper GNNs~\cite{DeCorr}. Besides the fairness issue by sensitive attributes, bias can also come from the node degree~\cite{tang2020investigating}, graph condensation\cite{jin2021graph}, or even class distribution~\cite{dpgnn}, which we leave for future investigations.

%% file: conclusion.tex
\vspace{-1.5ex}
\section{Conclusion}\label{sec-conclusion}
In this paper, we focus on alleviating discrimination in learned node representations and made predictions on graphs from the perspective of sensitive leakage to innocuous features. Specifically, we empirically observe a novel problem that feature propagation could vary feature correlation and further cause sensitive leakage to innocuous feature channels, which may exacerbate discrimination in predictions. To tackle this problem, we propose FairVGNN to automatically mask sensitive-correlated feature channels considering the effect of correlation variation after feature propagation and adaptively clamp weights of encoder to absorb less sensitive information. Experimental results demonstrate the effectiveness of the proposed FairVGNN framework in achieving better trade-off between utility and fairness than other baselines. Some interesting phenomena are also observed such as the variation of correlation depends on different datasets, and the group fairness is related to the network homophily. 
Thus, one future direction would be to theoretically analyze the relationships among feature propagation, network homophily and correlation variation. Furthermore, we plan to leverage self-supervised learning~\cite{wang2022graph,jin2020self} to constrain the bias encoded in the learned node representation, and consider fairness in multi-sensitive groups in future work.

\begin{table*}[h]
\scriptsize
\setlength{\extrarowheight}{.095pt}
\setlength\tabcolsep{3pt}
\centering
\caption{Detailed comparison with different weight regularization strategies.}
\vskip -2.25ex
\label{tab-fullwc}
\begin{tabular}{l|l|ccccc|ccccc|ccccc}
\hline
\multirow{2}{*}{\textbf{Model}} & \multirow{2}{*}{\textbf{Strategy}} & \multicolumn{5}{c|}{\textbf{German}} & \multicolumn{5}{c|}{\textbf{Credit}} & \multicolumn{5}{c}{\textbf{Bail}} \\
 &  & AUC $(\uparrow)$ & ACC $(\uparrow)$ & F1 $(\uparrow)$ & $\Delta_{\text{sp}} (\downarrow)$ & $\Delta_{\text{eo}} (\downarrow)$ & AUC $(\uparrow)$ & ACC $(\uparrow)$ & F1 $(\uparrow)$ & $\Delta_{\text{sp}} (\downarrow)$ & $\Delta_{\text{eo}} (\downarrow)$ & AUC $(\uparrow)$ & ACC $(\uparrow)$ & F1 $(\uparrow)$ & $\Delta_{\text{sp}} (\downarrow)$ & $\Delta_{\text{eo}} (\downarrow)$ \\
 \hline
\multirow{3}{*}{\textbf{GCN}} & Ad wc & 72.41+2.10 & 70.16+0.86 & 82.15+0.42 & 1.71+1.68 & 0.88+0.58 & 71.34+0.41 & 78.04+0.33 & 87.08+0.74 & 5.02+5.22 & 3.60+4.31 & 85.68+0.37 & 84.73+0.46 & 79.11+0.33 & 6.53+0.67 & 4.95+1.22 \\
 & Wc & 73.34+0.91 & 70.08+0.59 & 81.91+0.38 & 4.54+2.98 & 4.22+3.35 & 69.61+4.11 & 77.79+0.98 & 86.70+0.72 & 4.67+4.01 & 3.48+3.02 & 85.97+0.45 & 85.12+0.26 & 79.08+0.28 & 6.86+0.47 & 5.85+0.83 \\
 & Sn & 71.15+0.95 & 70.24+0.48 & 82.01+0.35 & 3.94+5.37 & 2.82+3.76 & 70.16+2.65 & 75.96+1.40 & 84.87+2.03 & 5.67+4.53 & 4.45+3.62 & 86.10+0.61 & 85.69+0.42 & 79.66+0.63 & 7.53+0.17 & 6.43+0.81 \\
 \hline
\multirow{3}{*}{\textbf{GIN}} & Ad wc & 71.65+1.90 & 70.16+0.32 & 82.40+0.14 & 0.43+0.54 & 0.34+0.41 & 74.05+0.42 & 80.16+0.19 & 88.10+0.30 & 5.09+1.30 & 2.67+0.92 & 83.22+1.60 & 83.86+1.57 & 76.36+2.20 & 5.67+0.76 & 5.77+1.26 \\
 & Wc & 71.62+2.32 & 71.04+0.60 & 82.76+0.22 & 1.82+0.82 & 0.48+0.40 & 73.20+1.20 & 79.03+1.09 & 87.23+0.94 & 7.03+4.58 & 4.74+3.47 & 84.67+0.67 & 84.79+1.45 & 78.43+1.11 & 7.73+0.44 & 6.96+1.26 \\
 & Sn & 71.25+1.69 & 72.40+1.10 & 82.98+0.68 & 7.73+5.03 & 2.37+1.42 & 71.12+0.55 & 78.54+2.00 & 86.53+1.90 & 2.60+0.73 & 0.87+0.54 & 85.47+0.74 & 85.26+1.28 & 79.13+1.08 & 7.07+1.72 & 5.90+2.06 \\
 \hline
\multirow{3}{*}{\textbf{SAGE}} & Ad wc & 73.84+0.52 & 70.00+0.25 & 81.91+0.63 & 1.36+1.90 & 1.22+1.49 & 74.05+0.20 & 79.94+0.30 & 87.84+0.32 & 4.94+1.10 & 2.39+0.71 & 91.56+1.71 & 88.41+1.29 & 83.58+1.88 & 1.14+0.67 & 1.69+1.13 \\
 & Wc & 72.43+1.60 & 70.48+0.85 & 82.03+0.82 & 4.85+4.10 & 2.50+2.12 & 73.20+1.21 & 79.03+1.09 & 87.23+0.94 & 7.03+4.58 & 4.74+3.47 & 91.48+0.57 & 88.42+0.78 & 83.51+0.88 & 3.45+1.06 & 1.89+1.27 \\
 & Sn & 73.00+1.53 & 70.00+1.07 & 81.82+0.59 & 3.74+3.22 & 1.89+1.08 & 73.86+0.54 & 78.57+1.70 & 86.55+1.67 & 5.27+3.26 & 3.78+2.51 & 93.36+1.75 & 89.88+1.07 & 85.32+1.71 & 2.55+1.19 & 1.52+1.27\\
 \hline
\end{tabular}
\begin{tablenotes}
  \footnotesize
  \item \textbf{*} \textbf{Ad wc}: adaptively clamp weights of the encoder;  \textbf{Wc}: clamp weights of the encoder; and \textbf{Sn}: spectral normalization of the encoder.
  
  
\end{tablenotes}
\vskip -2.25ex
\end{table*}

\vspace{-2ex}
\section{Acknowledgements}
\vspace{-0.5ex}
Yushun Dong and Jundong Li are supported by the National Science Foundation (NSF) under grant No. 2006844 and the Cisco Faculty Research Award.

%% file: appendix.tex
\section{Summary of Notations}\label{sec-notation}
To facilitate understanding, we present a summary of commonly utilized notations and the corresponding descriptions in Table~\ref{tb:symbols}.

\begin{table}[htbp!]
\setlength{\extrarowheight}{.095pt}
\setlength\tabcolsep{2pt}
\caption{Notations commonly used in this paper and the corresponding descriptions.} 
\vspace{-1ex}
\label{tb:symbols}
\begin{tabular}{cc}
\hline
\hline
\textbf{Notations}       & \textbf{Definitions or Descriptions} \\
\hline
\hline
$G$   &  input graph\\
$\mathcal{V}$, $\mathcal{E}$   &  node, edge set\\
$\mathbf{A}$   &  adjacency matrix\\
$\mathbf{X}$  &  node attribute matrix\\
$\mathbf{m}$ & feature mask\\
$\tau$ & temperature factor\\
$\chi$ & network homophily\\
$\epsilon$ & prefix cutting threshold\\
$\mathbf{S}$   &  sensitive feature vector\\
$\widetilde{\mathbf{X}}$  &  generated node attribute matrix\\
$\mathbf{Y}$   &  one-hot encoded label matrix for all nodes\\

$\Delta_{\text{sp}}, \Delta_{\text{eo}}$   &  statistical parity and equality of opportunity\\
$y, s$   &  class label and sensitive group label\\
$\rho_i$   &  the Pearson correlation coefficient of the $i^{\text{th}}$ channel\\
$\mathbf{H}^L$ & representation learned after $L$-layers GNNs\\
$\mathbf{W}^{f,1}$ & weight of the first layer of the encoder\\

$g, d, f, c$ & generator, discriminator, encoder and classifier\\

\hline
\hline
\end{tabular}
\vspace{1ex}
\end{table}

\vspace{-1ex}
\section{Experimental Settings}\label{app-experiment}
\subsection{Detailed Model Architecture}\label{app-architecture}
A unified template of a graph convolutional layer is formalized as:
\begin{equation}\label{eq-propagation}
    \mathbf{h}_i^l = \text{TRAN}^{l}(\text{PROP}^{l}(\mathbf{h}_{i}^{l - 1}, \{\mathbf{h}_j^{l - 1}|j\in\mathcal{N}_i\})),
\end{equation}
 where $\mathcal{N}_i$ denotes the neighborhood set of node $v_i$ and $\text{PROP}^l, \text{TRAN}^l$ stand for neighborhood propagation and feature transformation at layer $l$. In neighborhood propagation, neighborhood representations are propagated and further fused with itself to get the intermediate representation $\widehat{\mathbf{h}}_i^l$. 
Then, the $\text{TRAN}^l$ function is applied on $\widehat{\mathbf{h}}_i^l$ to get the final representation $\mathbf{h}_i^l$ of node $v_i$ at layer $l$. Note that $\mathbf{h}_i^0$ of node $v_i$ is typically initialized as the original node feature $\mathbf{X}_{i}$. After stacking $L$ graph convolutional layers, every node  aggregates their neighborhood information up to $L$-hops away and we denote it as $\mathbf{H}^L\in\mathbb{R}^{n\times d^{L}}$. 
 Many graph convolutions 
 can be obtained under this template by configuring different $\text{PROP}^{l}$ and $\text{TRAN}^{l}$. In this work, the encoder of FairVGNN is designed following this template. 

We use GCN, GIN and GraphSAGE as our GNN-backbones respectively for each bias-alleviating method. The basic graph convolution layer of these three backbones, respectively, are:
\begin{equation}\label{eq:GCN}
    \mathbf{H}^{l} = \widetilde{\mathbf{D}}^{-0.5}(\mathbf{A} + \mathbf{I})\widetilde{\mathbf{D}}^{-0.5}\mathbf{H}^{l - 1}\mathbf{W}^{l},
\end{equation}
\begin{equation}\label{eq:GIN}
    \mathbf{H}^{l} = \text{MLP}^l((\mathbf{A} + (1 + \alpha)\mathbf{I})\mathbf{H}^{l - 1}),
\end{equation}
\begin{equation}\label{eq:SAGE}
    \mathbf{H}^{l} = \mathbf{W}^{l, 1}\mathbf{H}^{l - 1} + \mathbf{W}^{l, 2}\mathbf{D}^{-1}\mathbf{A}\mathbf{H}^{l - 1},
\end{equation}
where $\widetilde{\mathbf{D}}$ is the degree matrix with added self-loop, $\mathbf{H}^{l - 1}$ is the node representation obtained from the previous layer and $\mathbf{H}^{0} = \mathbf{X}$. In this work, we only consider one graph convolution, therefore $l = 1$.

\vspace{-1ex}
\subsection{Hyperparameter for Each Baseline}\label{app-hyper}
As different bias-alleviating methods have different model architectures, their hyperparameters are also different and are presented respectively in the following:

\begin{itemize}[leftmargin=*]
    \item \textbf{NIFTY}: dropout \{0.0, 0.5, 0.8\}, the number of hidden unit 16, learning rate $\{1e^{-2}, 1e^{-3}, 1e^{-4}\}$, project hidden unit 16, weight decay $\{1e^{-4}, 1e^{-5}\}$, drop edge rate $0.001$, drop feature rate $0.1$, regularization coefficient $\{0.4, 0.5, 0.6, 0.7, 0.8\}$.
    
    \item \textbf{EDITS}: initial learning rate 0.003, weight decay $1e^{-7}$, threshold proportions for Credit, German, and Recidivism dataset are 0.02, 0.25, 0.012 respectively.
    
    \item \textbf{FairGNN}: dropout $\{0.0, 0.5, 0.8\}$, the number of hidden unit 32, learning rate $\{0.0001, 0.001, 0.01\}$, weight decay $1e^{-5}$, regularization coefficients $\alpha=4, \beta=0.01$, sensitive number $200$, label number $500$.
\end{itemize}

\vspace{-1ex}
\section{Dataset Details}\label{app-dataset}
Here we present the detailed description of three datasets we used to validate our proposed FairVGNN as follows:
\begin{itemize}[leftmargin=*]
    \item \textbf{German Credit} (German): nodes are clients in a German bank, node attributes include gender, loan amount, and other account-related details, with edges formed between clients if their credit accounts are similar. The task is to classify the credit risk of the clients as high or low with `gender' being the sensitive feature.
    \item \textbf{Recidivism} (Bail): nodes 
    are defendants released on bail during 1990-2009, and edges are formed between defendants if they share similar past criminal records and demographics. The task is to predict whether a defendant would be more likely to commit a violent or nonviolent crime once released on bail with `race' being the sensitive feature.
    \item \textbf{Credit Defaulter} (Credit): nodes 
    are credit card users, and edges are formed between users if their share similar pattern in purchases/payments. The task is to predict whether a user will default on credit card payment with `age' being the sensitive feature.
\end{itemize} 

\section{Detailed Experimental Results}
\subsection{Effect of Weight Clamping}\label{app-wceffect}
Table~\ref{tab-fullwc} reports the full results of comparing our proposed adaptive weight clamping with two other weight regularization approaches: weight clamping and spectral normalization. We can clearly see that generally our proposed adaptive weight clamping achieves better trade-off between utility and fairness. This is because adaptive weight clamping clamps weights more on sensitive-related features and hence minimally remove critical information beneficial for classification. However, in some cases such as on Credit dataset, the GCN with adaptive weight clamping has higher bias than directly weight clamping. This is because blindly clamping weights with no selection would remove more information, some of which might overlap with sensitive information and hence cause less bias, while some of which might overlap with class-related information and hence cause lower model utility (the accuracy is 77.79 lower than 78.04 when using adaptive weight clamping).

\subsection{Detailed Correlation Variation of German and Credit Datasets}\label{app-corrvalfull}
Here we visualize the correlation variation of the first 13 feature channels on German and Credit after different layers of feature propagation. We clearly see that compared with German where some feature channels quickly become highly correlated to sensitive channel while some become less correlated. The sensitive correlation of feature channels on Credit changes more slowly. Therefore, masking according to the rank of original sensitive correlation $\boldsymbol{\rho}^{\text{origin}}$ is roughly the same as the propagated sensitive correlation $\boldsymbol{\rho}^{\text{prop}}$ and the performance of S$_1$ and S$_2$ are the same in Table~\ref{tab-prelim} on Credit compared with German. We argue that the slow variance of feature correlation is because the higher homophily of Credit (0.9595) than German (0.8048) triggers less change of feature correlation during feature propagation. In the extreme case where node features strictly obey the network homophily, feature propagation would cause no change on feature distributions of every node and therefore the feature correlation would stay the same.

\begin{figure}[t]
     \centering
     \begin{subfigure}[b]{0.45\textwidth}
         \centering
         \hspace{-8.5ex}
         \includegraphics[width=0.93\textwidth]{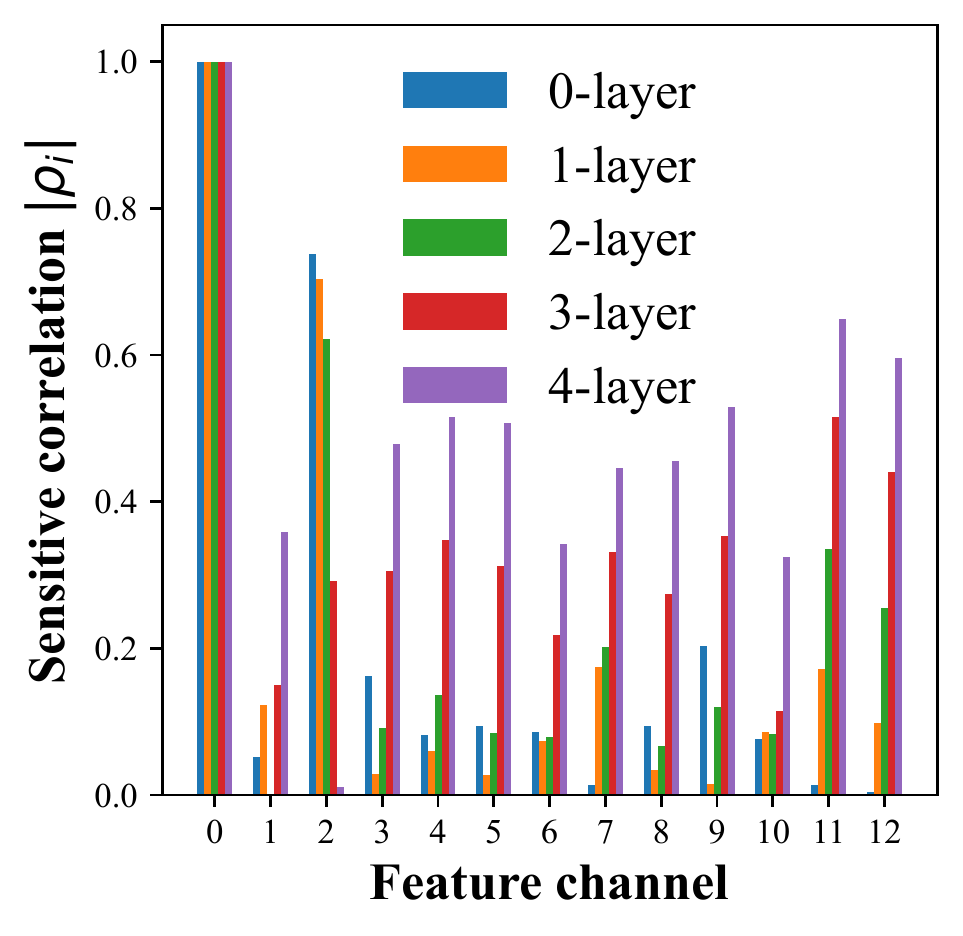}
         \vskip -1.5ex
         \caption{German (0.8048)}
         \label{fig-germancorrfull}
     \end{subfigure}
     \vskip 1.5ex
     \begin{subfigure}[b]{0.45\textwidth}
         \centering
         \hspace{-8.5ex}
         \includegraphics[width=0.93\textwidth]{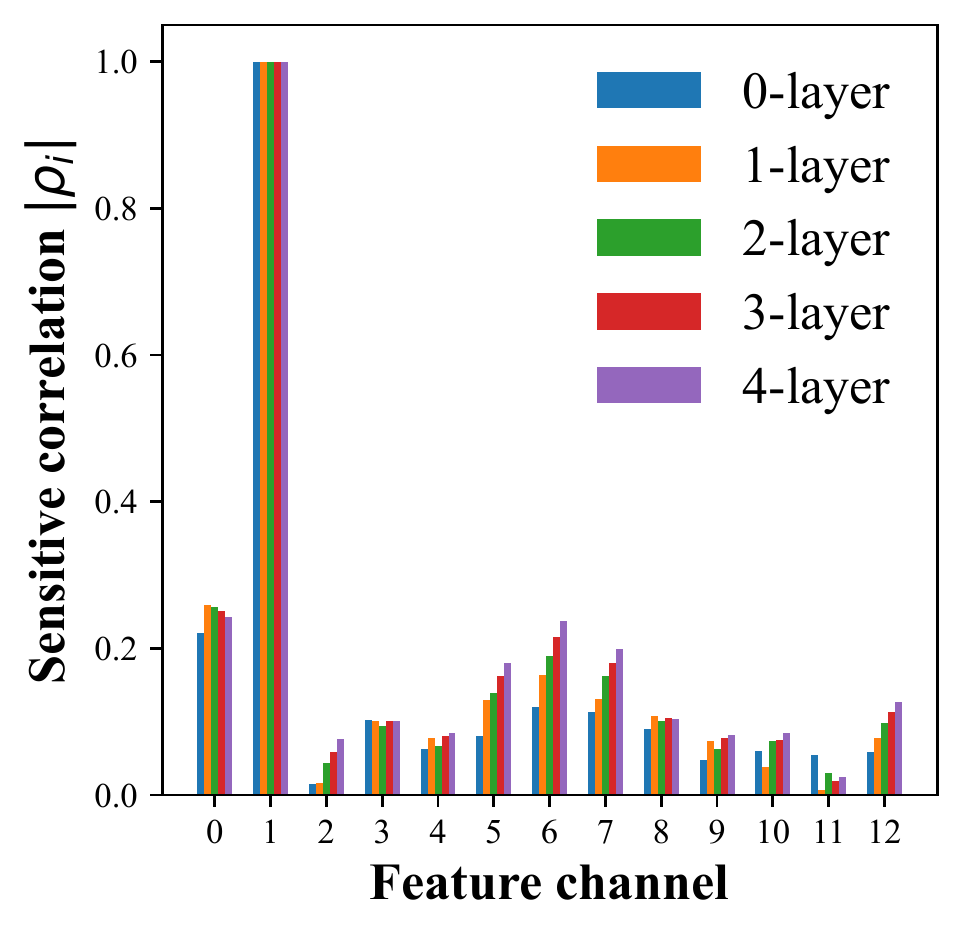}
         \vskip -1.5ex
         \caption{Credit (0.9595)}
         \label{fig-creditcorrfull}
     \end{subfigure}
     \vskip -1.5ex
     \caption{Correlation variation after feature propagation on the German and Credit datasets with the parentheses next to each dataset denoting the network homophily.}
     \label{fig-corrvar}
\end{figure}
